%% file: sam_dynamics_icml.tex
\documentclass[nohyperref]{article}

\usepackage{microtype}
\usepackage{graphicx}
\usepackage{subfigure}
\usepackage{booktabs} % for professional tables

\usepackage{hyperref}

\usepackage[accepted]{icml2023}

\usepackage{amsmath}
\usepackage{amssymb}
\usepackage{mathtools}
\usepackage{amsthm}

\input{math_commands.tex}

\usepackage[capitalize,noabbrev]{cleveref}

\theoremstyle{plain}
\newtheorem{theorem}{Theorem}[section]

\theoremstyle{definition}

\theoremstyle{remark}

\makeatletter
\newtheorem*{rep@theorem}{\rep@title}
\newcommand{\newreptheorem}[2]{%
\newenvironment{rep#1}[1]{%
 \def\rep@title{#2 \ref{##1}}%
 \begin{rep@theorem}}%
 {\end{rep@theorem}}}
\makeatother

\newreptheorem{theorem}{Theorem}
\newreptheorem{corollary}{Corollary}
\newreptheorem{lemma}{Lemma}
\usepackage[textsize=tiny]{todonotes}

\usepackage{mathmacros}

\newcommand{\Lo}{\mathcal{L}}
\newcommand{\w}{\m{w}}
\renewcommand{\v}{\m{v}}
\renewcommand{\H}{\m{H}}
\newcommand{\rad}{\rho}
\newcommand{\al}{\alpha}
\newcommand{\bt}{\beta}
\newcommand{\diag}{{\rm diag}}

\renewcommand{\lr}{\eta}

\newcommand{\lam}{\lambda}
\newcommand{\J}{\m{J}}
\newcommand{\z}{\m{z}}
\newcommand{\G}{\m{G}}
\newcommand{\Q}{\tens{Q}}

\renewcommand{\th}{\sm{\theta}}
\newcommand{\D}{D}
\renewcommand{\P}{P}
\newcommand{\B}{B}
\newcommand{\x}{\m{x}}
\newcommand{\y}{\m{y}}

\newcommand{\f}{\m{f}}

\newcommand{\sing}{\sigma}

\newcommand{\zz}{z}

\newcommand{\tl}{\tilde}
\renewcommand{\tJ}{\tl{\J}}
\newcommand{\tz}{\tl{\z}}

\newcommand{\rat}{r}
\newcommand{\bfr}{\beta}
\newcommand{\pmat}{\m{P}}
\newcommand{\gm}{\gamma}

\newif\ifcomments
\commentsfalse % uncomment to turn all colored comments off

\ifcomments
\newcommand{\ynd}[1]{{\color{blue}[YD: #1]}}
\newcommand{\aga}[1]{{\color{red}[AA: #1]}}
\newcommand{\hmb}[1]{{\color{green}[HM: #1]}}
\else
\newcommand{\ynd}[1]{}
\newcommand{\aga}[1]{}
\newcommand{\hmb}[1]{}
\fi

\icmltitlerunning{SAM operates far from home}

\begin{document}

\twocolumn[
\icmltitle{SAM operates far from home: eigenvalue regularization as a dynamical phenomenon\aga{comments on}}



\icmlsetsymbol{equal}{*}

\begin{icmlauthorlist}
\icmlauthor{Atish Agarwala}{equal,goog}
\icmlauthor{Yann Dauphin}{equal,goog}
\end{icmlauthorlist}

\icmlaffiliation{goog}{Brain Team, Google Research}

\icmlcorrespondingauthor{Atish Agarwala}{thetish@google.com}

\icmlkeywords{Machine Learning, ICML}

\vskip 0.3in
]



\printAffiliationsAndNotice{\icmlEqualContribution} % otherwise use the standard text.

\begin{abstract}
The Sharpness Aware Minimization (SAM) optimization algorithm has been shown to control large eigenvalues of the loss Hessian and provide generalization benefits in a variety of settings. The original motivation for SAM was a modified loss function which penalized sharp minima; subsequent analyses have also focused on the behavior near minima. However, our work reveals that SAM provides a strong regularization of the eigenvalues throughout the learning trajectory. We show that in a simplified setting, SAM dynamically induces a stabilization related to the edge of stability (EOS) phenomenon observed in large learning rate gradient descent. Our theory predicts the largest eigenvalue as a function of the learning rate and SAM radius parameters. Finally, we show that practical models can also exhibit this EOS stabilization, and that understanding SAM must account for these dynamics far away from any minima.
\end{abstract}

\section{Introduction}

\label{sec:introduction}

Since the dawn of optimization, much effort has gone into developing algorithms which
use geometric information about the loss landscape to make optimization more
efficient and stable
\cite{nocedal_updating_1980, duchi_adaptive_2011, lewis_nonsmooth_2013}. In more modern machine
learning,
control of the large curvature eigenvalues of the loss landscape has been
a goal in and of itself
\cite{hochreiter_flat_1997, chaudhari_entropysgd_2019}. There is empirical
and theoretical evidence that
controlling curvature of the training landscape leads to benefits for generalization
\cite{keskar_largebatch_2017, neyshabur_exploring_2017},
although in general the relationship between the two is
complex \cite{dinh_sharp_2017}.

Recently the \emph{sharpness aware minimization} (SAM)
algorithm has emerged as a popular choice for
regularizing the curvature during training \cite{foret_sharpnessaware_2022}. SAM has the
advantage of being a tractable first-order method; for the cost of a single extra
gradient evaluation, SAM can control the large eigenvalues of the loss Hessian
and often leads to improved optimization and generalization \cite{bahri_sharpnessaware_2022}.

However, understanding the mechanisms behind the effectiveness of SAM is an open question. The SAM algorithm itself
is a first-order approximation of SGD on a modified loss function
$\tl{\Lo}(\th) = \max_{||\delta\th||<\rad}\Lo(\th+\delta\th)$. Part of the original
motivation was that $\tl{\Lo}$ explicitly penalizes sharp minima over flatter ones. However the approximation
performs as well or better than running gradient descent on $\tl{\Lo}$ directly.
% , even though the first-order method does not correspond to gradient
% descent on any loss function
% \aga{are we the first to notice this? If so we should make it clear in the text}\ynd{I believe so, but I think we should remove this observation for now. It would require further explanation as it stands}.
SAM often works better with small batch sizes as compared to larger ones
\cite{foret_sharpnessaware_2022, andriushchenko_understanding_2022}.%\aga{other refs?}
These stochastic effects suggest that studying the deterministic gradient flow dynamics on $\tl{\Lo}$
will not capture key features of SAM, since small batch size induces non-trivial
differences from gradient flow \cite{paquette_sgd_2021}.

In parallel to the development of SAM, experimental and theoretical work has uncovered
some of the curvature-controlling properties of first-order methods due to finite step
size - particularly in the full batch setting. At intermediate learning rates, a wide
variety of models and optimizers show a tendency for the largest Hessian eigenvalues
to stabilize near the \emph{edge of stability} (EOS) for long times
\cite{lewkowycz_large_2020, cohen_gradient_2022, cohen_adaptive_2022}. The EOS is the
largest eigenvalue which would lead to convergence for a quadratic loss landscape.
This effect can be explained in terms of a non-linear feedback between the
large eigenvalue and changes in the parameters in that eigendirection
\cite{ damian_selfstabilization_2022, agarwala_secondorder_2022}.

We will show that these two areas of research are in fact intimately linked: under
a variety of conditions, SAM displays a modified EOS behavior, which leads to stabilization
of the largest eigenvalues at a lower magnitude via non-linear, discrete dynamics.
These effects highlight the dynamical nature of eigenvalue regularization, and demonstrates
that SAM can have strong effects throughout a training trajectory.

\subsection{Related work}

Previous experimental work suggested that decreasing batch size causes
SAM to display
both stronger regularization and better generalization
\cite{andriushchenko_understanding_2022}. This analysis also suggested
that SAM may induce more sparsity.

A recent theoretical approach studied SAM close to a minimum, where the trajectory
oscillates about the minima and provably decreases
the largest eigenvalue \cite{bartlett_dynamics_2022}. A contemporaneous approach
studied the SAM algorithm in the limit of small learning rate and SAM radius, and quantified
how the implicit and explicit regularization of SAM differs between full batch and
batch size $1$ dynamics \cite{wen_how_2023}.

\subsection{Our contributions}

\hmb{Maybe slightly rewords the opening of this section, to say the high level take-home message of this study is that SAM's trajectory and behavior intricately relates to earlier stages of training, and analysis only near convergence cannot capture the full picture. Then you can say, For characterizing the following bullet points as the technical pieces of your story that when put together, reveal the important role of the trajectory in SAM's behavior.}

In contrast to other theoretical approaches, we study the behavior of SAM far from minima. We find
that SAM regularizes the eigenvalues throughout training through a dynamical phenomenon and analysis
only near convergence cannot capture the full picture. In particular, in simplified models we
show:
% In contrast to other theoretical approaches, we study SAM far from
% minima, and in a regime which is non-perturbative in both learning rate
% and the SAM radius.
% Using simplified models, we show the following:
\begin{itemize}
\setlength\itemsep{0em}
\item Near initialization, full batch SAM provides limited suppression of large eigenvalues (Theorem
\ref{thm:j_dyn_gd}).
\item SAM induces a modified edge of stability (EOS) (Theorem \ref{thm:eos_SAM}).
\item For full batch training, the largest eigenvalues stabilize at the SAM-EOS, at a smaller
value than pure gradient descent (Section \ref{sec:basic_experiments}).
\item As batch size decreases, the effect of SAM is stronger and the
dynamics is no longer controlled by the Hessian alone (Theorem \ref{thm:zj_dyn_sgd}).
\end{itemize}
We then present experimental results on realistic models which show:
\begin{itemize}
\item The SAM-EOS predicts the largest eigenvalue for WideResnet 28-10 on CIFAR10. 
\end{itemize}
Taken together, our results suggest that SAM can operate throughout the learning trajectory,
far from minima,
and that it can use non-linear, discrete dynamical effects to stabilize
large curvatures of the loss function.

\section{Quadratic regression model}

\label{sec:quad_model}

\subsection{Basic model}

We consider a \emph{quadratic regression model} \cite{agarwala_secondorder_2022}
which extends a linear regression
model to second order in the parameters. Given a $\P$-dimensional parameter vector $\th$,
the $\D$-dimensional output is given by $\f(\th)$:
\begin{equation}
\f(\th) = \y+\G^\top\th+\frac{1}{2}\Q(\th, \th)\,.
\end{equation}
Here, $\y$ is a $\D$-dimensional vector,
$\G$ is a $\D\times\P$-dimensional matrix, and $\Q$ is a $\D\times\P\times\P$-
dimensional
tensor symmetric in the last two indices - that is, $\Q(\cdot, \cdot)$ takes two
$\P$-dimensional vectors as input, and outputs a $\D$-dimensional vector $\Q(\th, \th)_\alpha = \th^\top \Q_{\alpha} \th$.
If $\Q = \boldsymbol{0}$, the model corresponds to linear regression. $\y$, $\G$, and $\Q$
are all fixed at initialization.

Consider optimizing the model with under a squared loss. More
concretely, let $\y_{tr}$ be a $\D$-dimensional vector of
training targets. We focus on the MSE loss
\begin{equation}
\Lo(\th) = \frac{1}{2}||\f(\th)-\y_{tr}||^{2}
\end{equation}

We can write the dynamics in terms of the residuals $\z$ and the Jacobian
$\J$ defined by
\begin{equation}
\z \equiv f(\th)-\y_{tr},~\J \equiv \frac{\partial\f}{\partial\th} = \G+\Q(\th,\cdot)\,.
\end{equation}
The loss can be written as $\Lo(\th) = \frac{1}{2}\z\cdot\z$.
The full batch gradient descent (GD) dynamics of the parameters are given by
\begin{equation}
\th_{t+1} = \th_{t}-\lr\J^{\top}_{t}\z_{t}
\end{equation}
which leads to
\begin{equation}
\begin{split}
\z_{t+1}-\z_{t} & = -\lr\J_{t} \J_{t}^{\top}\z_{t}  +\frac{1}{2}\lr^2 \Q(\J_{t}^{\top}\z_{t},\J_{t}^{\top}\z_{t})\\
\J_{t+1} -\J_{t} & = -\lr \Q(\J_{t}^{\top}\z_{t}, \cdot)\,.
\label{eq:GD_ingeneral}
\end{split}
\end{equation}
The $\D\times\D$-dimensional matrix $\J\J^{\top}$ is known as the
\emph{neural tangent kernel} (NTK) \cite{jacot_neural_2018},
and controls the dynamics for small $\lr||\J^{\top}\z||$ \cite{lee_wide_2019}.

We now consider the dynamics of un-normalized SAM \cite{andriushchenko_understanding_2022}.
That is, given a loss function
$\Lo$ we study the update rule
\begin{equation}
\th_{t+1}-\th_{t} = -\lr\nabla\Lo(\th_{t}+\rad\nabla\Lo(\th_{t}))
\label{eq:unnorm_sam}
\end{equation}
We are particularly interested
in small learning rate and small SAM radius. The dynamics in $\z-\J$ space
are given by
\begin{equation}
\begin{split}
\z_{t+1}-\z_{t} & = -\lr\J\J^{\top}(1+\rad\J\J^{\top})\z-\lr\rad \z\cdot\Q(\J^{\top}\z, \J^{\top}\cdot)\\
&+\lr^2\frac{1}{2}\Q(\J^{\top}\z, \J^{\top}\z)+O(\lr\rad(\lr+\rad)||\z||^{2})
\end{split}
\label{eq:z_dyn_expansion}
\end{equation}
\begin{equation}
\begin{split}
\J_{t+1}-\J_{t} & = -\lr\left[\Q((1+\rad\J^{\top}\J)\J^{\top}\z, \cdot)+\right.\\
&\left.\rad\Q(\z\cdot\Q(\J^{\top}\z, \cdot), \cdot)\right]+O(\lr\rad^2||\z||^{2})
\end{split}
\label{eq:J_dyn_expansion}
\end{equation}
to lowest order in $\lr$ and $\rad$.

From Equation \ref{eq:z_dyn_expansion} we see that for small $\lr||\z||$ and $\rad||\z||$,
the dynamics of $\z$ is controlled by the modified NTK $(1+\rad\J\J^{\top})\J\J^{\top}$.
The factor $1+\rad\J\J^{\top}$ shows up in the dynamics of $\J$ as well, and we will show
that this effective NTK can lead to dynamical stabilization of large eigenvalues. And note that when $\rho=0$, these dynamics coincide with that of gradient descent.

\subsection{Gradient descent theory}

\subsubsection{Eigenvalue dynamics at initialization}

A basic question is: how does SAM affect the eigenvalues of the NTK?
We can study this directly for early learning
dynamics by using random initializations. We have the following theorem
(proof in Appendix \ref{app:quad_average}):
\begin{theorem}
\label{thm:j_dyn_gd}
Consider a second-order regression model, with $\Q$ initialized randomly with
i.i.d. components with $0$ mean and variance $1$.
For a model trained with full batch gradient descent, with unnormalized SAM,
the change in $\J$ at the
first step of the dynamics, averaged over $\Q$ is
\begin{equation}
\expect_{\Q}[\J_{1} -\J_{0}]  = -\rad\lr\P\z_{0}\z_{0}^{\top}\J_{0}+O(\rad^2\lr^2||\z_{0}||^2)+O(\lr^3||\z_{0}||^3)
\end{equation}
The $\al$th
singular value $\sing_{\al}$ of $\J_{0}$ associated
with left and right singular vectors 
$\w_{\al}$ and $\v_{\al}$ can be approximated as
\begin{equation}
\begin{split}
&(\sing_{\al})_{1}-(\sing_{\al})_{0}  =
\w_{\al}^{\top}\expect_{\Q}[\J_{1} -\J_{0}]\v_{\al} +O(\lr^2)\\
& = -\rad\lr\P (\z_{0}\cdot\w_{\al})^{2}\sing_{\al}+O(\lr^2)
\end{split}
\end{equation}
for small $\lr$.
\end{theorem}
Note that the singular vector $\w_{\al}$ is an eigenvector of $\J\J^{\tpose}$
associated with the eigenvalue $\sing_{\al}^{2}$.

This analysis suggests that on average, at early times, the change in the
singular value is negative. However, the change also depends linearly
on $(\w_{\al}\cdot\z_{0})^{2}$. This suggests that if the component of $\z$
in the direction of the singular vector becomes small, the stabilizing effect of SAM
becomes small as well. For large batch size/small learning rate with MSE loss,
we in fact expect $\z\cdot\w_{\al}$ to decrease rapidly early in training
\cite{cohen_gradient_2022, agarwala_secondorder_2022}.
Therefore the relative regularizing effect can be \emph{weaker} for larger
modes in the GD setting.

\begin{figure}[h]
    \centering
    \includegraphics[width=0.8\linewidth]{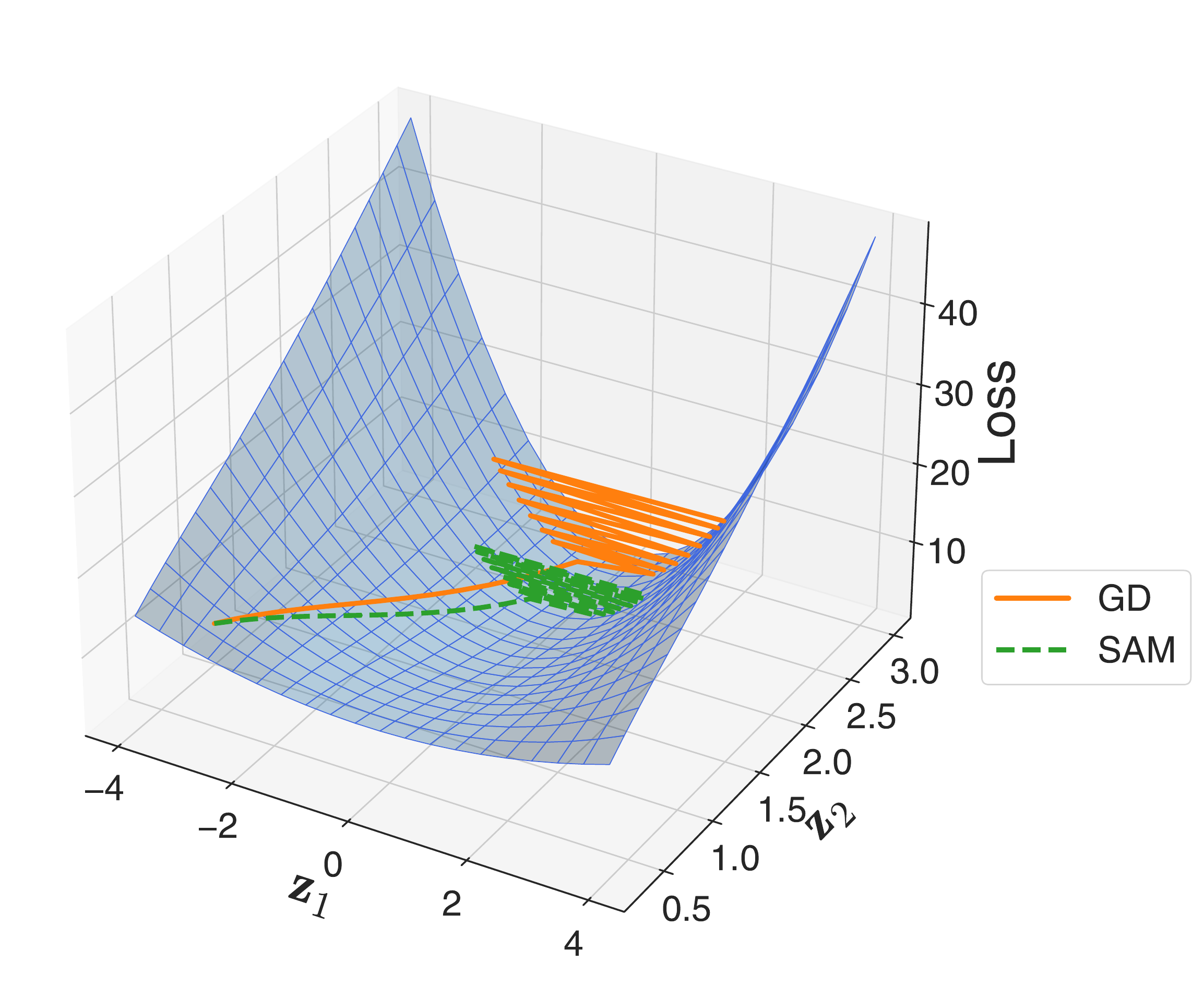}
    \caption{Schematic of SAM-modified EOS. Gradient descent decreases loss
    until a high-curvature area is reached, where large eigenmode is non-linearly
    stabilized (orange, solid). SAM causes stabilization to happen earlier, at a smaller
    value of the curvature (green, dashed).}
    \label{fig:eos_schematic}
\end{figure}

\begin{figure*}[t]
    \centering
    \begin{tabular}{c c c}
         \includegraphics[width=0.3\linewidth]{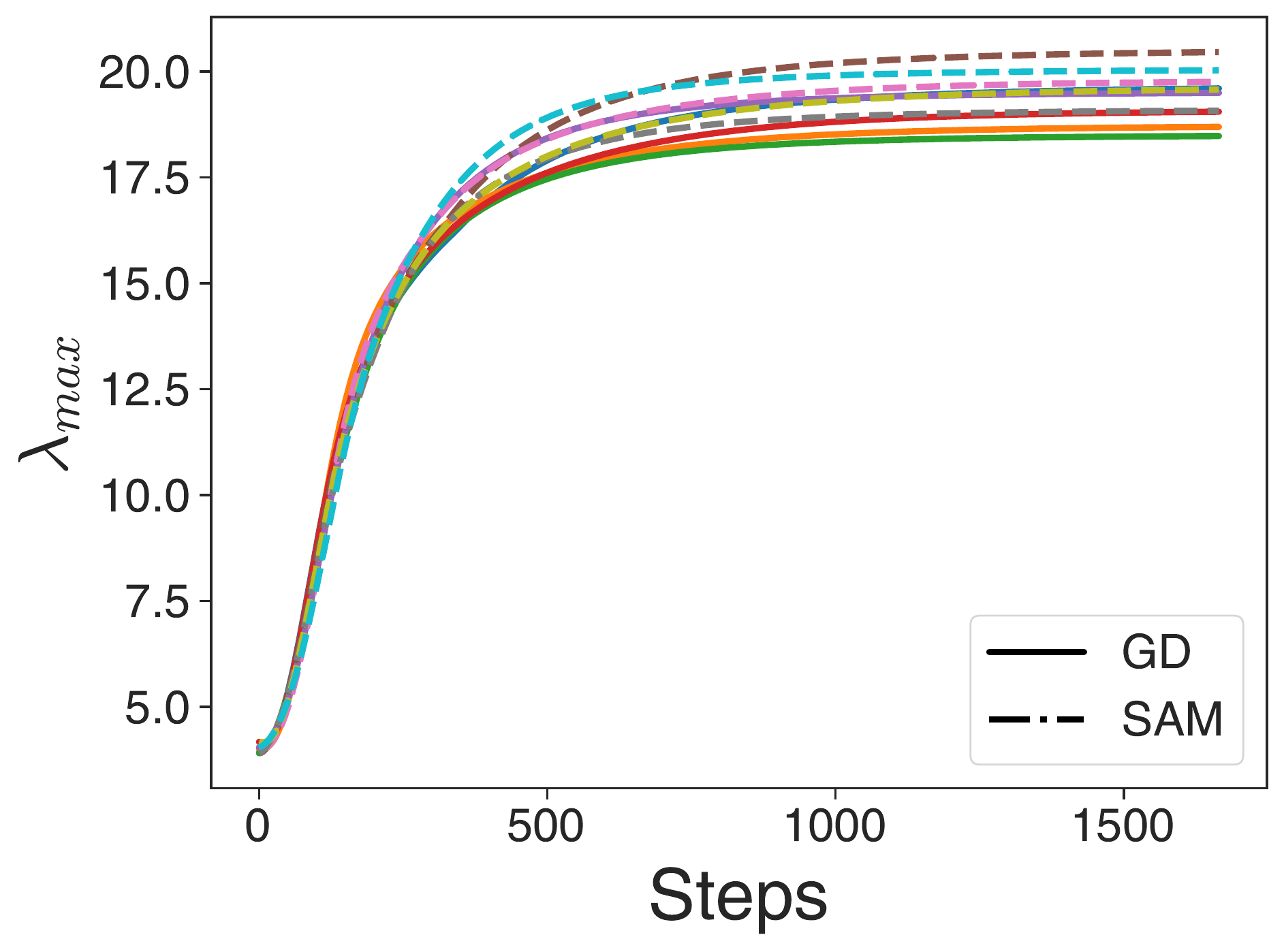} &  
         \includegraphics[width=0.3\linewidth]{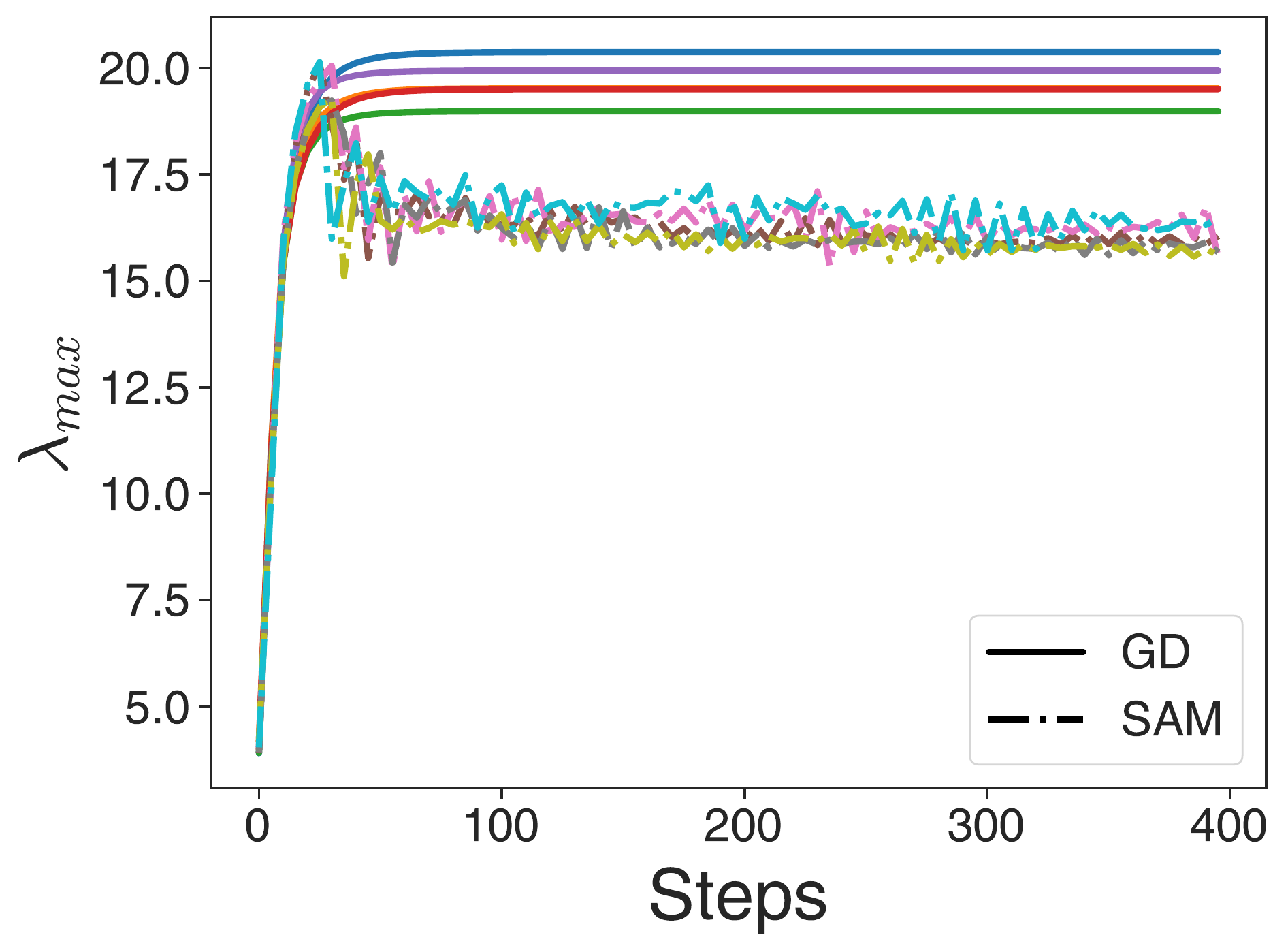} & 
         \includegraphics[width=0.3\linewidth]{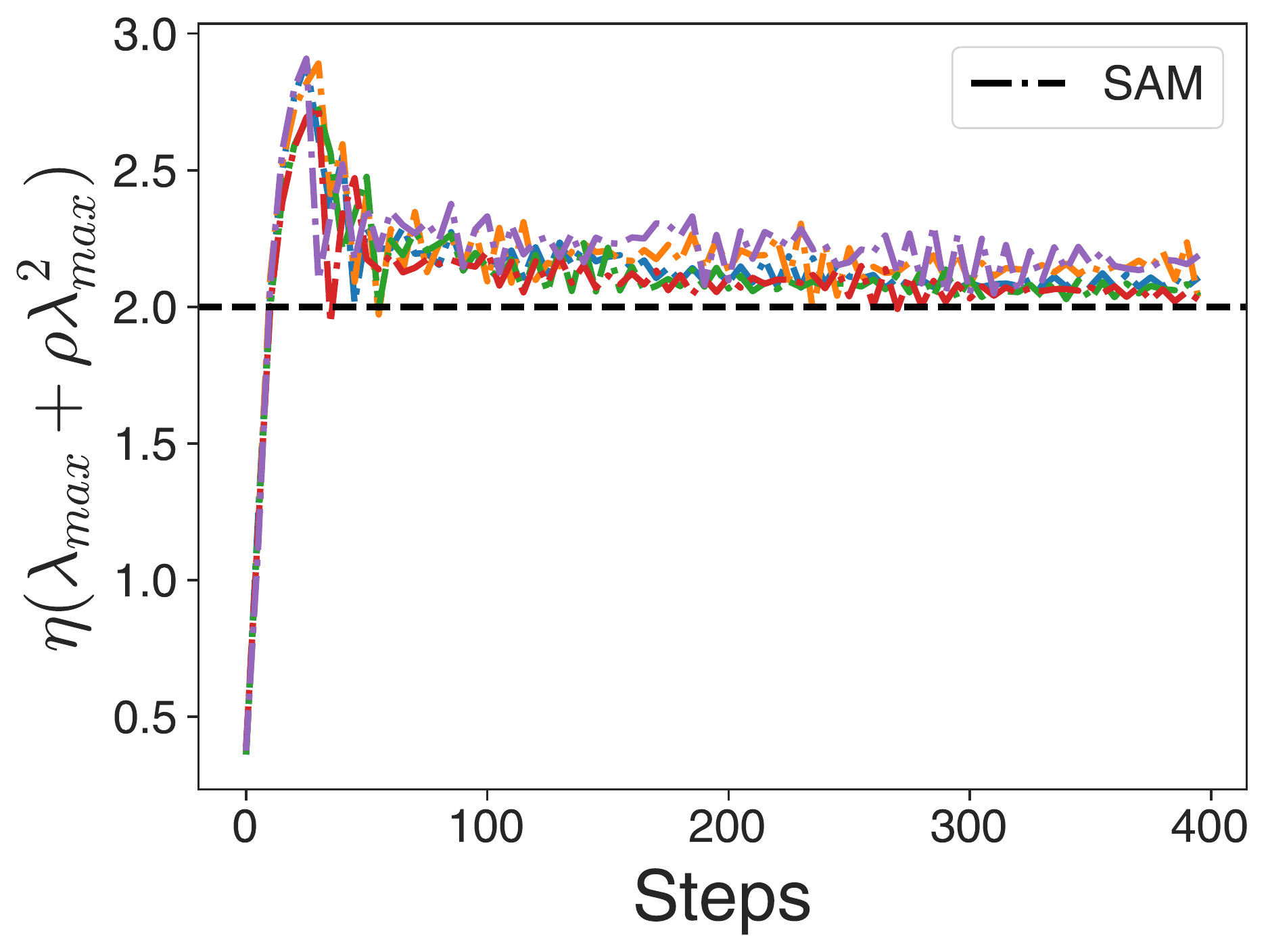}
    \end{tabular}
    \caption{Trajectories of largest eigenvalue $\lam_{max}$ of $\J\J^{\top}$
    for quadratic regression model, $5$ independent initializations.
    For gradient descent with small learning rate ($\lr = 3\cdot10^{-3}$),
    SAM ($\rad = 4\cdot10^{-2}$) does not regularize the large NTK eigenvalues (left).
    For larger learning rate ($\lr = 8\cdot10^{-2}$), SAM controls large eigenvalues (middle).
    Largest eigenvalue can be predicted by SAM edge of stability $\lr(\lam_{max}+\rad\lam_{max}^{2}) = 2$ (right).}
    \label{fig:sam_quad_model}
\end{figure*}

\subsubsection{Edge of stability and SAM}

One of the most dramatic consequences of SAM for full batch training
is the shift of the \emph{edge of stability}.
We begin by reviewing the EOS phenomenology.
Consider full-batch gradient descent
training with respect to a twice-differentiable loss. Near a minimum of the loss,
the dynamics of the displacement $\x$ from the minimum (in parameter space)
are well-approximated by
\begin{equation}
\x_{t+1}-\x_{t} = -\lr\H\x_{t}
\end{equation}
where $\H$ is the positive semi-definite Hessian at the minimum
$\x = 0$. The dynamics converges
exponentially iff the largest eigenvalue of $\H$ is bounded by
$\lr\lam_{max} < 2$. We refer to $\lr\lam_{max}$ as the
\emph{normalized eigenvalue},
Otherwise, there is at least one component
of $\x$ which is non-decreasing. The value $2/\lr$ is often referred to as the
\emph{edge of stability} (EOS) for the dynamics.

Previous work has shown that for many non-linear models,
there is a range of learning rates
where the largest eigenvalue of the Hessian stabilizes
around the edge of stability \cite{cohen_gradient_2022}.
Equivalent phenomenology exists for other gradient-based methods \cite{cohen_adaptive_2022}.
The stabilization effect is due to feedback between the largest curvature eigenvalue and
the displacement in the largest eigendirection \cite{agarwala_secondorder_2022, damian_selfstabilization_2022}. For MSE loss, EOS behavior occurs for the large
NTK eigenvalues as well \cite{agarwala_secondorder_2022}.

We will show that SAM also induces
an EOS stabilization effect, but at a smaller eigenvalue than GD.
We can understand the shift intuitively by analyzing
un-normalized SAM on a loss $\frac{1}{2}\x^{\top}\H\x$. Direct
calculation gives the update rule:
\begin{equation}
\x_{t+1}-\x_{t} = -\lr(\H+\rad\H^{2})\x_{t}
\label{eq:low_order_dyn}
\end{equation}
For positive definite $\H$, $\x_{t}$ converges exponentially to $0$
iff $\lr(\lam_{max}+\rad\lam_{max}^2)<2$. 
Recall from Section 2.1 that the SAM NTK is $(1 +\rho \J\J^{\top})\J\J^{\top}> \J\J^{\top}$.
This suggests that
$\lr(\lam_{max}+\rad\lam_{max}^2)$ is the \emph{SAM normalized eigenvalue}.
This bound gives a critical $\lam_{max}$ which is smaller than that
in the GD case. This leads to the hypothesis that SAM can cause
a stabilization at the EOS in a flatter region of the loss, as schematically
illustrated in Figure \ref{fig:eos_schematic}.

We can formalize the \emph{SAM edge of stability} (SAM EOS) for any differentiable model
trained on MSE loss. Equation
\ref{eq:z_dyn_expansion} suggests the matrix
$\J\J^{\top}(1+\rad\J\J^{\top})$ - which has larger eigenvalues for
larger $\rad$ - controls the low-order
dynamics. We can formalize this intuition in the following theorem
(proof in Appendix \ref{app:sam_eos_proof}):
\begin{theorem}
\label{thm:eos_SAM}
Consider a $\mathcal{C}^{\infty}$ model $\f(\th)$
trained using Equation \ref{eq:unnorm_sam} with MSE loss. Suppose that there
exists a point $\th^*$ where $\z(\th^*) = 0$. Suppose that for some
$\eps>0$, we have the lower bound
$\eps < \lr\lam_{i}(1+\rad\lam_{i})$ for the eigenvalues of the
positive definite symmetric matrix
$\J(\th^*)\J(\th^*)^{\top}$. Given a bound on the largest eigenvalue,
there are two regimes:

\textbf{Convergent regime.} If $\lr\lam_{i}(1+\rad\lam_{i})<2-\eps$ for all
for all eigenvalues $\lam_{i}$ of $\J(\th^*)\J(\th^*)^{\top}$, there exists a neighborhood
$U$ of $\th^*$ such that $\lim_{t\to\infty}\z_{t} = 0$ with exponential convergence for
any trajectory initialized at $\th_{0}\in U$.

% \textbf{Divergent regime.} If $\lr\lam_{i}(1+\rad\lam_{i})>2+\eps$ for some eigenvector
% $\v_{i}$ of $\J(\th^*)\J(\th^*)^{\top}$, then there exists some $(q_{min}, d_{min})$
% such that for any $q<q_{min}$, $d<d_{min}$, given $S_{q, d}$, the connected neighborhood
% of $\th^*$ such that $||\th-\th^*||<q$ and $||\z(\th)||<d$ for all $\th$ in $S_{q, d}$.
% there exists some initialization $\th_{0}\in S_{q, d}$ such that the trajectory
% $\{\th_{t}\}$ leaves $S_{q,d}$ at some finite time $t$.
\textbf{Divergent regime.} If $\lr\lam_{i}(1+\rad\lam_{i})>2+\eps$ for some eigenvector
$\v_{i}$ of $\J(\th^*)\J(\th^*)^{\top}$, then there exists some $q_{min}$
such that for any $q<q_{min}$, given $B_{q}(\th^*)$, the ball of radius $q$
around $\th^*$,
there exists some initialization $\th_{0}\in B_{q}(\th^*)$ such that the trajectory
$\{\th_{t}\}$ leaves $B_{q}(\th^*)$ at some time $t$.
\end{theorem}
Note that the theorem is proven for the NTK eigenvalues, which also show
EOS behavior for MSE loss in the GD setting \cite{agarwala_secondorder_2022}.

This theorem gives us the modified edge of stability
condition:
\begin{equation}
\lr \lam_{max}(1+\rad\lam_{max}) \approx 2
\label{eq:eos_sam_con}
\end{equation}
For larger $\rad$, a smaller $\lam_{max}$ is needed to meet the edge
of stability condition.
In terms of the
normalized eigenvalue $\tl{\lam} = \lr\lam$, the modified EOS can be written as
$\tl{\lam}(1+r\tl{\lam}) = 2$ with the ratio $\rat = \rad/\lr$.
Larger values of $\rat$ lead to stronger regularization effects, and
for the quadratic regression model specifically
$\lr$ can be factored out leaving $\rat$ as
the key dimensionless parameter (Appendix \ref{app:rescaled_dynamics}).

\subsection{SGD theory}

It has been noted that the effects of SAM have a strong dependence on
batch size \cite{andriushchenko_understanding_2022}. While a full analysis of
SGD is beyond the scope of this work, we can see
some evidence of stronger regularization for SGD in the quadratic regression
model.

Consider SGD dynamics, where a random fraction $\bfr = \B/\D$ of the training
residuals $\z$ are used to generate the dynamics at each step. We can represent
the sampling at each step with a random projection matrix $\pmat_{t}$, and replacing
all instances of $\z_{t}$ with $\pmat_{t}\z_{t}$. Under these dynamics, we can 
can prove the following:

\begin{theorem}
\label{thm:zj_dyn_sgd}
Consider a second-order regression model, with $\Q$ initialized randomly with
i.i.d. components with $0$ mean and variance $1$.
For a model trained with SGD, sampling $\B$ datapoints independently
at each step, the change in $\z$ and $\J$ at the first step, averaged
over $\Q$ and the sampling matrix $\pmat_{t}$, is given by
\begin{equation}
\begin{split}
&\expect[\z_{1}-\z_{0}]_{\Q,\pmat}   = -\lr\bfr\J_{0}\J_{0}^{\top}(1+\rad[\bfr(\J_{0}\J_{0}^{\top})\\
&+(1-\bfr)\diag(\J_{0}\J_{0}^{\top})])\z_{0}+O(\lr^2||\z||^{2})+O(\D^{-1})
\end{split}
\end{equation}
\begin{equation}
\begin{split}
& \expect_{\Q,\pmat}[\J_{1} -\J_{0}] = -\rad\lr\P (\bfr^{2}\z_{0}\z_{0}^{\top}+\bfr(1-\bfr)\diag(\z_{0}\z_{0}^{\top})) \J_{0}\\
&+O(\rad^2\lr^2||\z||^2)+O(\lr^3||\z||^{3})
\end{split}
\end{equation}
where $\bfr \equiv\B/\D$ is the batch fraction.
\end{theorem}
The calculations are detailed in Appendix \ref{app:quad_average}.
This suggests that there are two possible sources of increased regularization for
SGD: the first being the additional terms proportional to $\bfr(1-\bfr)$.
In addition to the fact that $\bfr(1-\bfr)>\bfr^2$ for $\bfr<\frac{1}{2}$, we have
\begin{equation}
\v_{\al}\diag(\z_{0}\z_{0}^{\top})) \J_{0}\w_{\al} = \sing_{\al} (\v_{\al}\circ\z_{0})\cdot(\v_{\al}\circ\z_{0})
\end{equation}
for left and right eigenvectors $\v_{\al}$ and $\w_{\al}$ of $\J_{0}$, where
$\circ$ is the Hadamard (elementwise) product. This term can be large even if
$\v_{\al}$ and $\z_{t}$ have small dot product.
This is in contrast to
$\bfr^2(\v_{\al}\cdot\z_{0})^{2}$, which is small if $\z_{0}$ does not have a large
component in the $\v_{\al}$ direction. This suggests that at short times, where
the large eigenmodes decay quickly, the SGD term can still be large.
Additionally, the onto the
largest eigenmode itself decreases more
slowly in the SGD setting \cite{paquette_sgd_2021}, which also suggests stronger
early time regularization for small batch size.

\begin{figure}[t]
\centering
\includegraphics[height=0.7\linewidth]{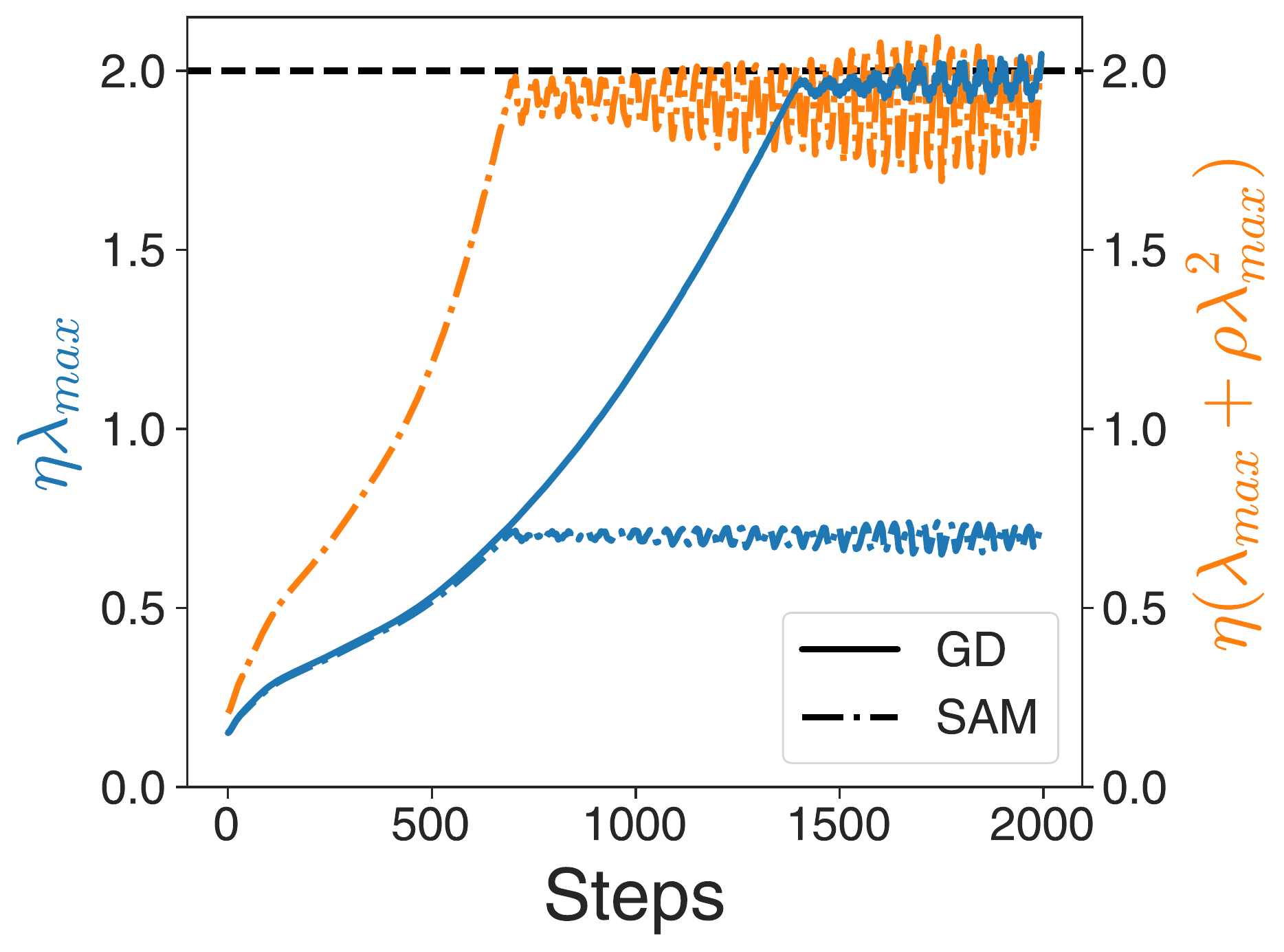}
\caption{Largest eigenvalues of $\J\J^{\top}$ for a fully-connected network trained using
MSE loss on 2-class CIFAR. For gradient descent ($\lr = 4\cdot10^{-3}$) largest eigenvalue
stabilizes according to the GD EOS $\lr\lam_{max} = 2$ (solid line, blue). SAM ($\rad = 10^{-2}$) 
stabilizes to a
lower value (dashed line, blue), which is well-predicted by the SAM EOS
$\lr(\lam_{max}+\rad\lam_{max}^2) = 2$ (dashed line, orange).}
\label{fig:cifar2_eigs}
\end{figure}

\section{Experiments on basic models}

\label{sec:basic_experiments}

\subsection{Quadratic regression model}

We can explore the effects of SAM and show the SAM EOS behavior
via numerical experiments on the quadratic
regression model. We use the update rule in Equation \ref{eq:unnorm_sam}, working
directly in $\z$ and $\J$ space as in \cite{agarwala_secondorder_2022}. Experimental details can be found in
Appendix \ref{app:quad_model_numerics}.

For small learning rates, we
see that SAM does not reduce the large eigenvalues of $\J\J^{\top}$ in the dynamics
(Figure \ref{fig:sam_quad_model}, left). In fact in some cases the final eigenvalue
is \emph{larger} with SAM turned on. The projection onto
the largest eigenmodes of $\J\J^{\top}$ exponentially decreases to $0$ quicker than
any other mode; as suggested by Theorem \ref{thm:j_dyn_gd}, this leads to only a small
decreasing pressure from SAM. The primary dynamics of the large eigenvalues is due
to the progressive sharpening phenomenology studied in
\cite{agarwala_secondorder_2022}, which tends to increase
the eigenmodes.

However, for larger learning rates, SAM has a strong suppressing effect on the
largest eigenvalues (Figure \ref{fig:sam_quad_model}, middle). The overall dynamics are
more non-linear than in the small learning rate case. The eigenvalues stabilize at the
modified EOS boundary $\lr(\lam_{max}+\rad\lam_{max}^2) = 2$
(Figure \ref{fig:sam_quad_model}, right), suggesting non-linear stabilization
of the eigenvalues. In Appendix \ref{app:quad_model_numerics}
we conduct additional experiments which confirm that the boundary predicts
the largest eigenvalue for a range of $\rad$, and that consequently
generally increasing $\rad$
leads to decreased $\lam_{max}$.

\subsection{CIFAR-$2$ with MSE loss}

We can see this phenomenology in more general non-linear models as well. We trained a
fully-connected network on the first $2$ classes of CIFAR with MSE loss, with both
full batch gradient descent and SAM.
We then computed the largest eigenvalues of $\J\J^{\top}$ along the trajectory. We
can see that in both GD and SAM the largest eigenvalues stabilize, and the stabilization
threshold is smaller for SAM (Figure \ref{fig:cifar2_eigs}). The threshold is once again well predicted by
the SAM EOS.

\section{Connection to realistic models}

\begin{figure*}[t]
    \centering
    \begin{tabular}{ccc}
        \includegraphics[width=0.31\linewidth]{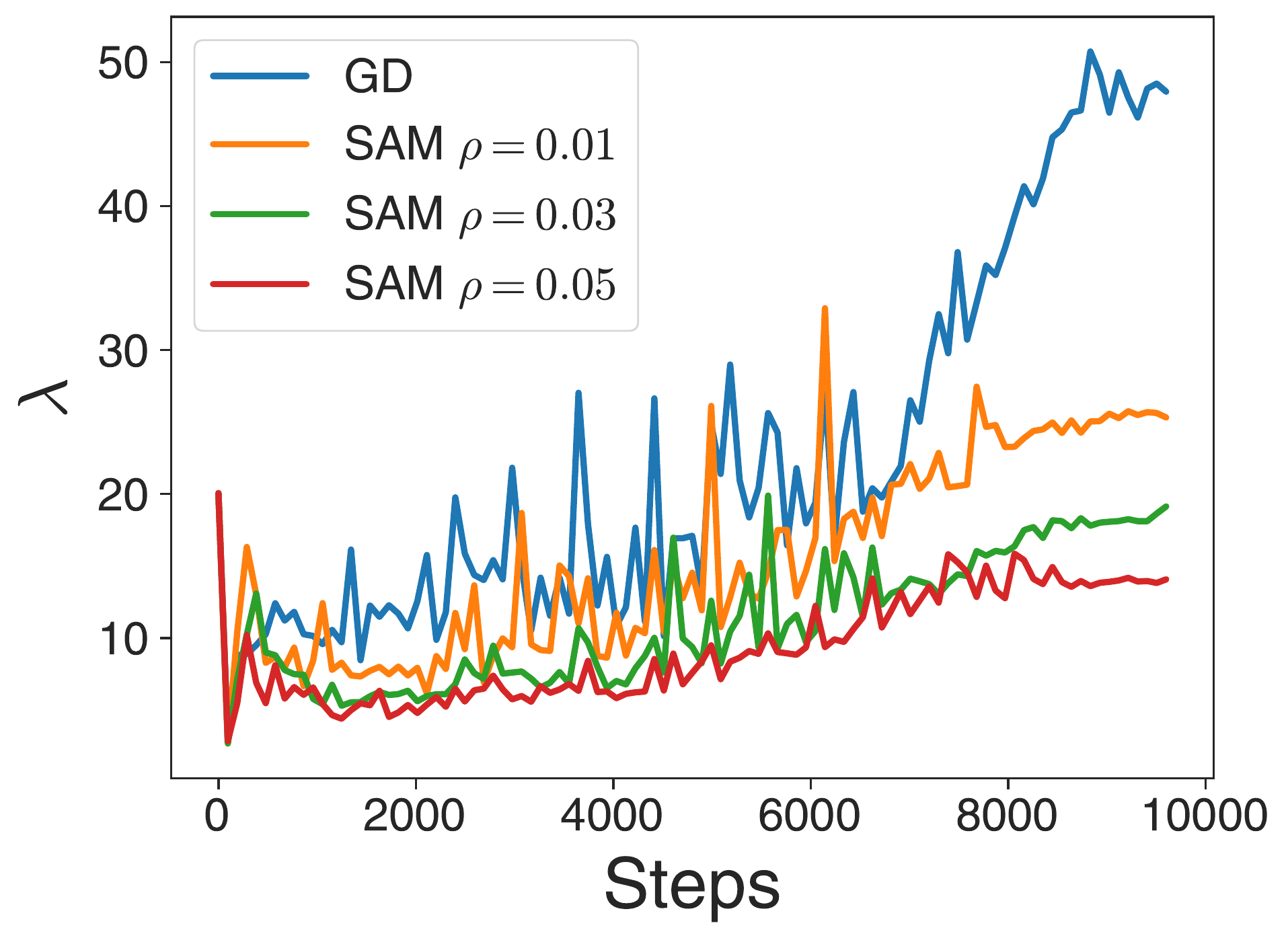}     & 
    \includegraphics[width=0.31\linewidth]{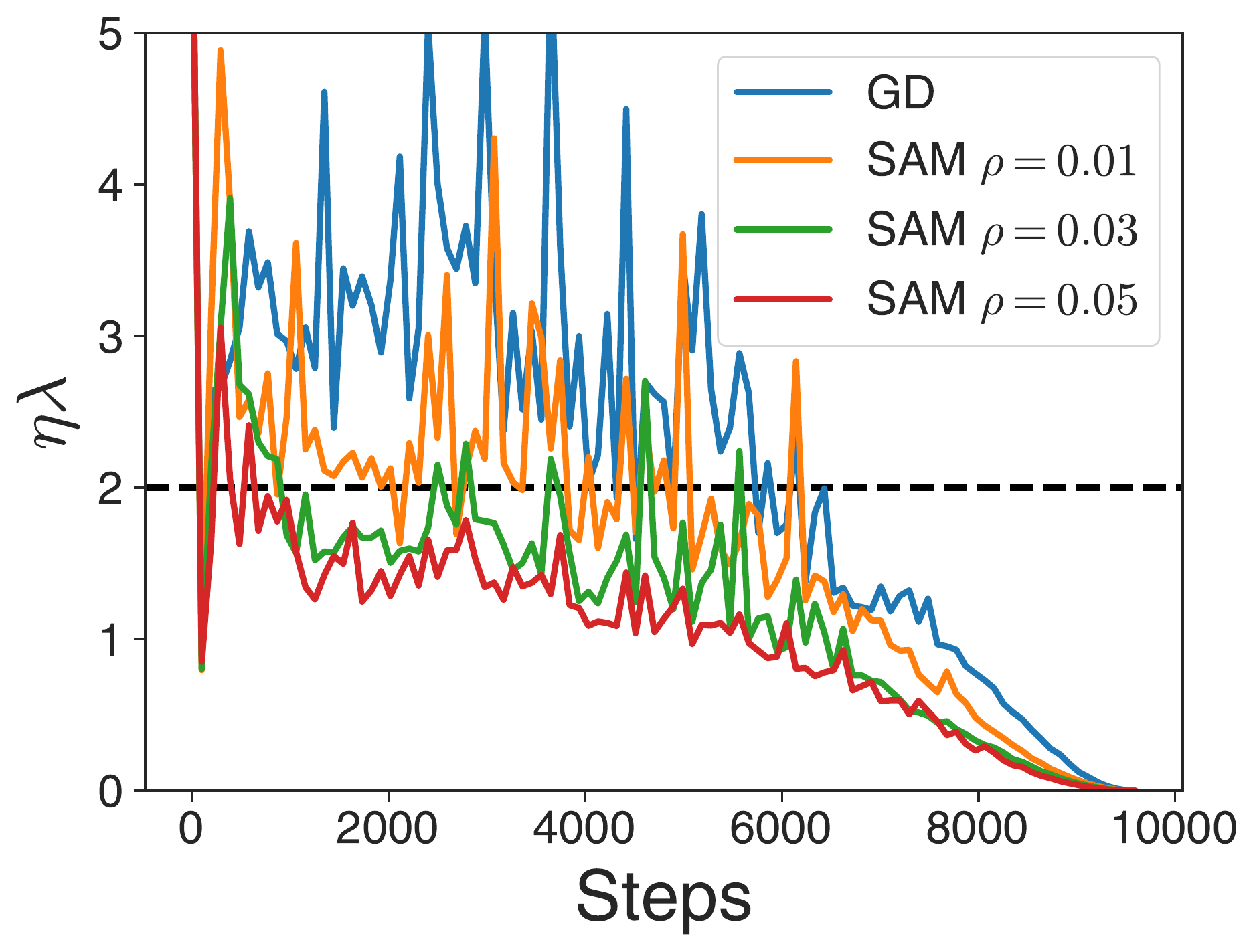}     & 
    \includegraphics[width=0.31\linewidth]{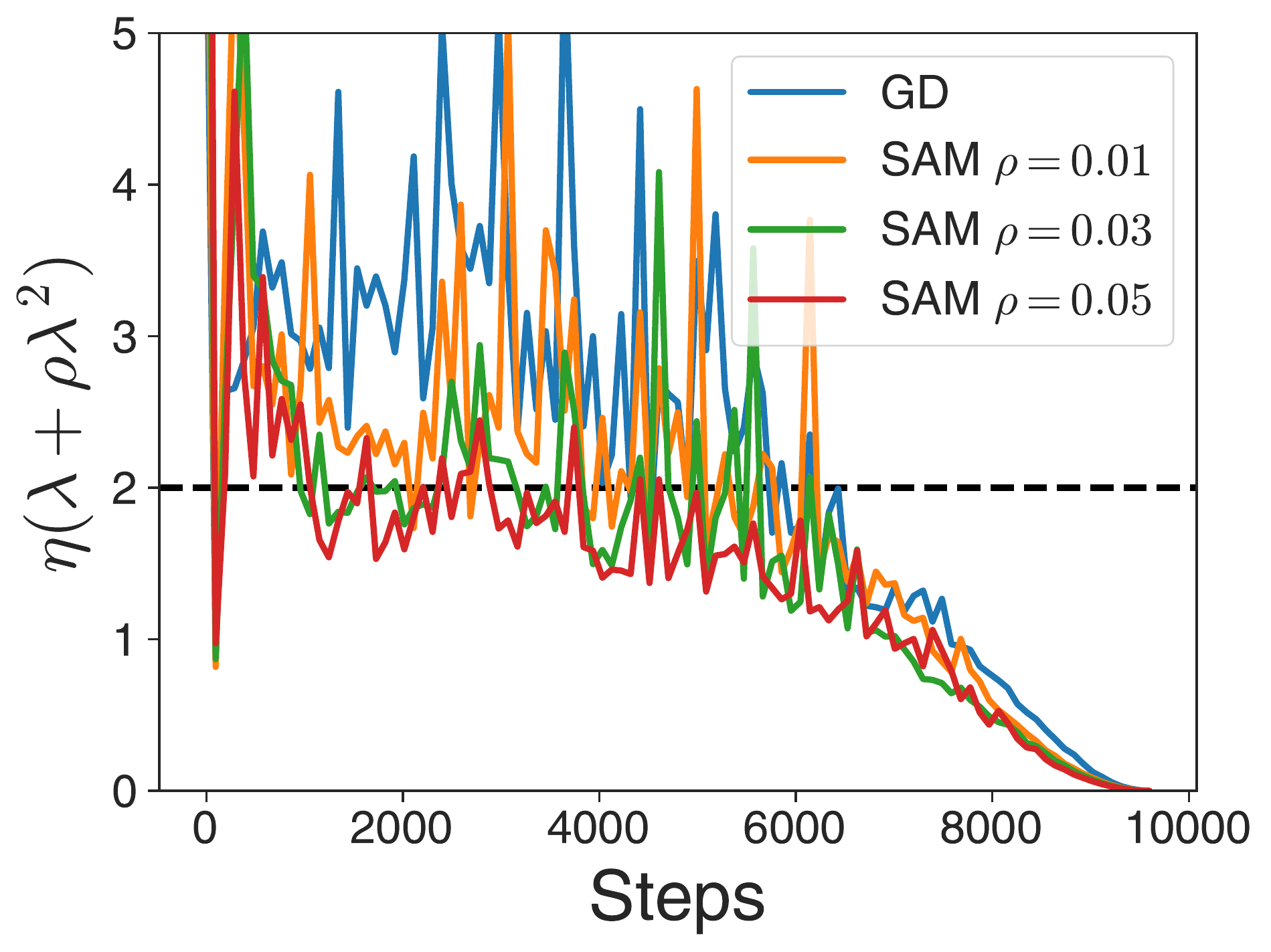}     
    \end{tabular}
    \caption{Largest Hessian eigenvalues for CIFAR10 trained with MSE loss. Left: largest eigenvalues
    increase at late times. Larger SAM radius mitigates eigenvalue increase.
    Middle: eigenvalues normalized by learning rate decrease at late times,
    and SGD shows edge of stability (EOS) behavior. Right: For larger $\rad$, SAM-normalized eigenvalues show modified EOS behavior.}
    \label{fig:cifar10_mse_eos}
\end{figure*}

\label{sec:experiments}

In this section, we show that our analysis of quadratic models can bring insights
into the behavior of more realistic models.

\subsection{Setup}

{\bf Sharpness} For MSE loss, edge of stability dynamics can be shown in terms of either the NTK eigenvalues
\emph{or} the Hessian eigenvalues \cite{agarwala_secondorder_2022}. For more general loss
functions, EOS dynamics takes place with respect to the largest Hessian eigenvalues
\cite{cohen_gradient_2022, damian_selfstabilization_2022}.
Following these results and the analysis in Equation \ref{eq:low_order_dyn}, we chose
to measure the largest eigenvalue of the Hessian rather than the NTK.
We used a Lanczos method \cite{ghorbani_investigation_2019} to approximately compute
$\lam_{max}$.
Any reference to $\lam_{max}$ in this section refers to
eigenvalues computed in this way.

{\bf CIFAR-10} We conducted experiments on the popular CIFAR-10 dataset \citep{krizhevsky2009learning} using the WideResnet 28-10 architecture \citep{zagoruyko2016wide}. We report results for both the MSE loss and the cross-entropy loss. In the case of the MSE loss, we replace the softmax non-linearity with Tanh and rescale the one-hot labels ${\bf y}\in \{0,1\}$ to $\{-1, 1\}$. In both cases, the loss is averaged across the number of elements in the batch and the number of classes. For each setting, we report results for a single configuration of the learning rate $\eta$ and weight decay $\mu$ found from an initial cross-validation sweep. For MSE, we use $\eta=0.3, \mu=0.005$ and $\eta=0.4, \mu=0.005$ for cross-entropy. We use the cosine learning rate schedule \citep{loshchilov2016sgdr} and SGD instead of Nesterov momentum \citep{sutskever2013importance} to better match the theoretical setup. Despite the changes to the optimizer and the loss, the test error for the models remains in a reasonable range (4.4\% for SAM regularized models with MSE and 5.3\% with SGD). In accordance with the theory, we use unnormalized SAM in these experiments.
We keep all other hyper-parameters to the default values described in the original WideResnet paper.

\subsection{Results}

% We measured the largest eigenvalue of the Hessian every $2$ epochs. 
As shown in Figure \ref{fig:cifar10_mse_eos} (left), the maximum eigenvalue increases significantly throughout training for all approaches considered.
However, the normalized curvature
$\lr\lam_{max}$, which sets the edge of stability in GD,
remains relatively stable early on
in training when the learning rate is high, but necessarily decreases as the cosine schedule drives the
learning rate to $0$ (Figure \ref{fig:cifar10_mse_eos}, middle).
% We can find a range of learning rates where SGD training
% shows EOS behavior; one such configuration is the one shown in Figure \ref{fig:cifar10_mse_eos}.

{\bf SAM radius drives curvature below GD EOS.}\quad As we increase the SAM radius, the largest 
eigenvalue is more controlled (Figure \ref{fig:cifar10_mse_eos}, left) - falling below the
gradient descent
edge of stability (Figure \ref{fig:cifar10_mse_eos}, middle). The stabilizing effect
of SAM on the large eigenvalues is evident even early on in training.

{\bf Eigenvalues stabilize around SAM-EOS.}\quad If we instead plot the SAM-normalized eigenvalue
$\lr(\lam_{max}+\rad\lam_{max}^{2})$, we see that the eigenvalues stay close to
(and often slightly above) the critical value of $2$, as predicted by theory
(Figure \ref{fig:cifar10_mse_eos}, right).
This suggests that there are settings where the control that SAM has
on the large eigenvalues of the Hessian comes, in part, from a modified
EOS stabilization effect.

\begin{figure*}[t]
    \centering
    \begin{tabular}{cc}
    \includegraphics[height=0.35\linewidth]{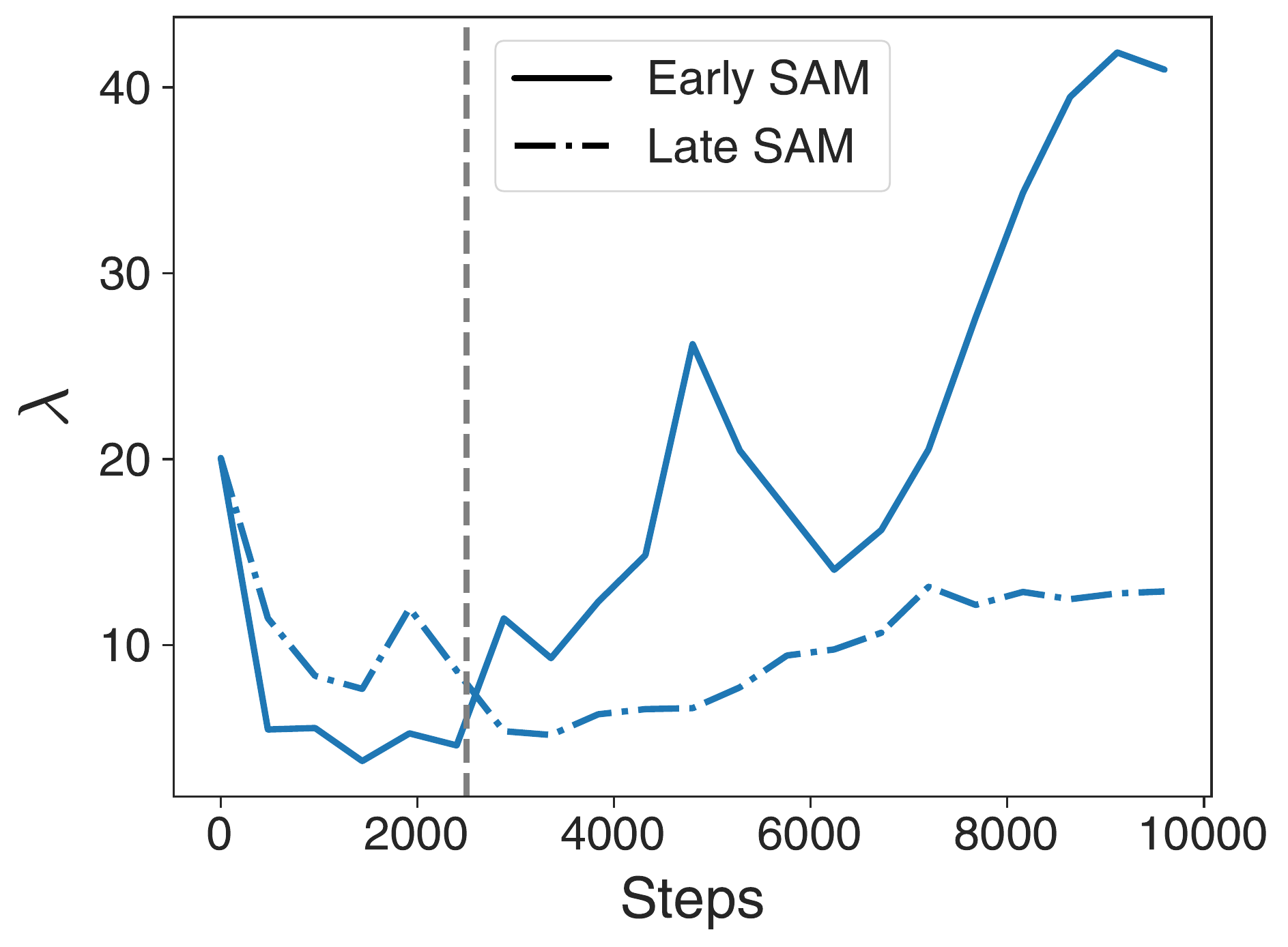} &
        \includegraphics[height=0.35\linewidth]{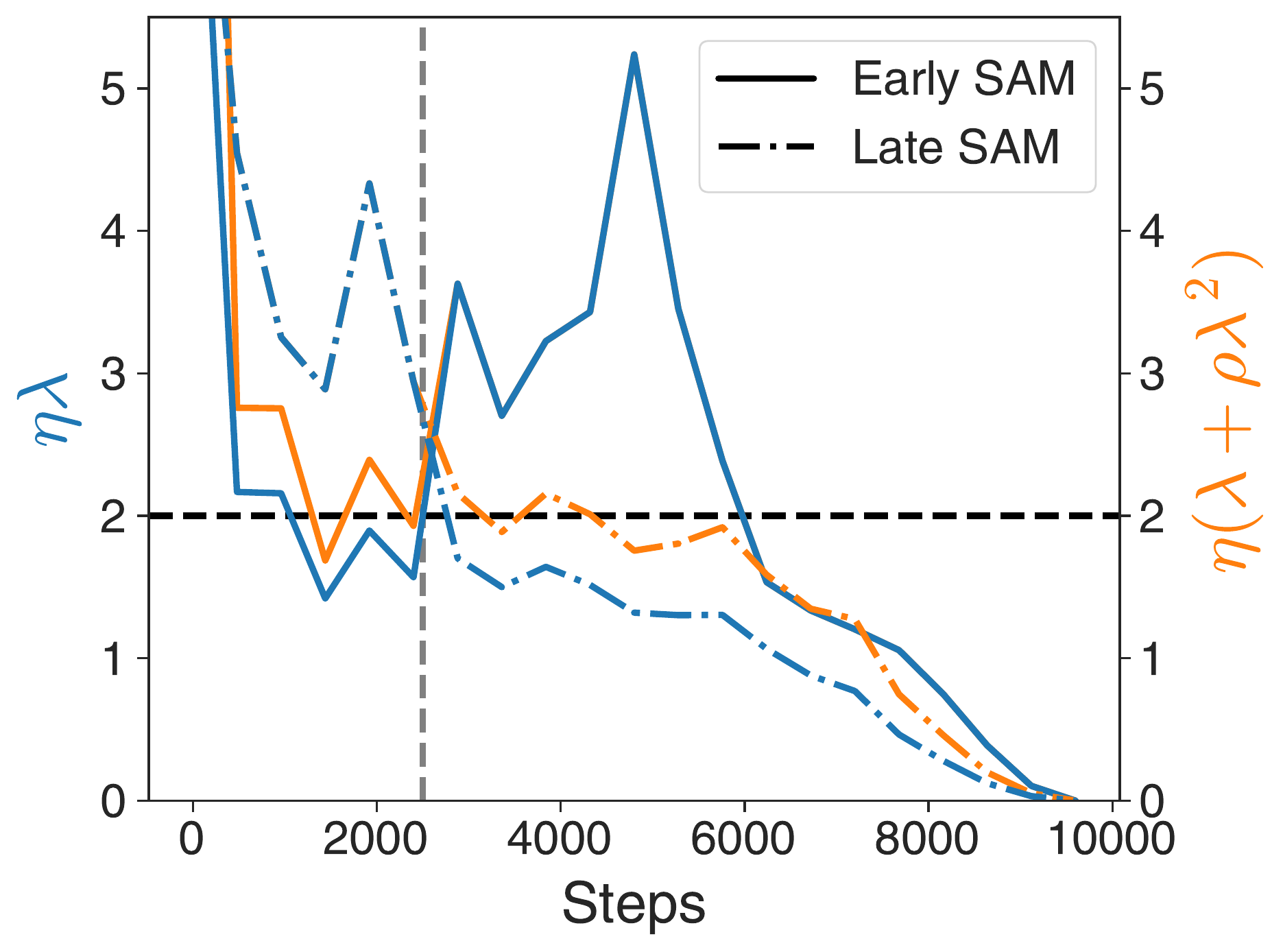}   
         
    \end{tabular}
    
    \caption{Maximum eigenvalues for CIFAR-10 model trained on MSE loss with a SAM
    schedule. Starting out with SAM ($\rad = 0.05$, solid lines) and turning it off at $2500$
    steps leads to initial suppression and eventual increase of $\lam_{max}$; starting out with SGD
    and turning SAM on after $2500$ steps leads to the opposite behavior (left). Eigenvalues
    cross over quickly after the switch. Plotting GD normalized eigenvalues (blue, right) shows
    GD EOS behavior in SGD phase; plotting SAM normalized eigenvalues (orange, right) shows SAM
    EOS behavior in SAM phase.}
    \label{fig:sam_sched}
\end{figure*}

{\bf Altering SAM radius during training can successfully move us between GD-EOS and SAM-EOS.}\quad Further evidence from EOS stabilization comes from using a \emph{SAM schedule}. We trained
the model with two settings: early SAM, where SAM is used for the first $2500$ steps ($50$ epochs),
after which the training proceeds with SGD ($\rho=0$), and late SAM, where SAM is used for the first
$2500$ steps, after which only SGD is used. The maximum eigenvalue is lower
for early SAM before $2500$ steps, after which there is a quick crossover and
late SAM gives better control (Figure \ref{fig:sam_sched}).
Both SAM schedules give improvement over SGD-only training. Generally, turning SAM on later or for the
full trajectory gave better generalization than turning SAM on early, consistent with the earlier
work of \citet{andriushchenko_understanding_2022}.

Plotting the eigenvalues for the early SAM and late SAM schedules, we see that when SAM is turned
off, the normalized eigenvalues lie above the gradient descent EOS (Figure \ref{fig:sam_sched},
right, blue curves). However when SAM is turned on, $\lr\lam_{max}$ is usually below the edge of stability
value of $2$; instead, the SAM-normalized value $\lr(\lam_{max}+\rad\lam_{max}^{2})$ lies
close to the critical value of $2$ (Figure \ref{fig:sam_sched}, right,
orange curves). This suggests that turning SAM on or off during the intermediate part of training
causes the dynamics to quickly reach the appropriate edge of stability.

{\bf Networks with cross-entropy loss behave similarly.}\quad We found similar results for cross-entropy loss as well, which we detail in Appendix 
\ref{app:x_ent_loss}.
The mini-batch gradient magnitude and eigenvalues vary more over the learning trajectories;
this may be related to effects of logit magnitudes which have been previously shown to affect
curvature and general training dynamics \cite{agarwala_temperature_2020, cohen_gradient_2022}.

\begin{figure}[ht]
    \centering
    \includegraphics[height=0.6\linewidth]{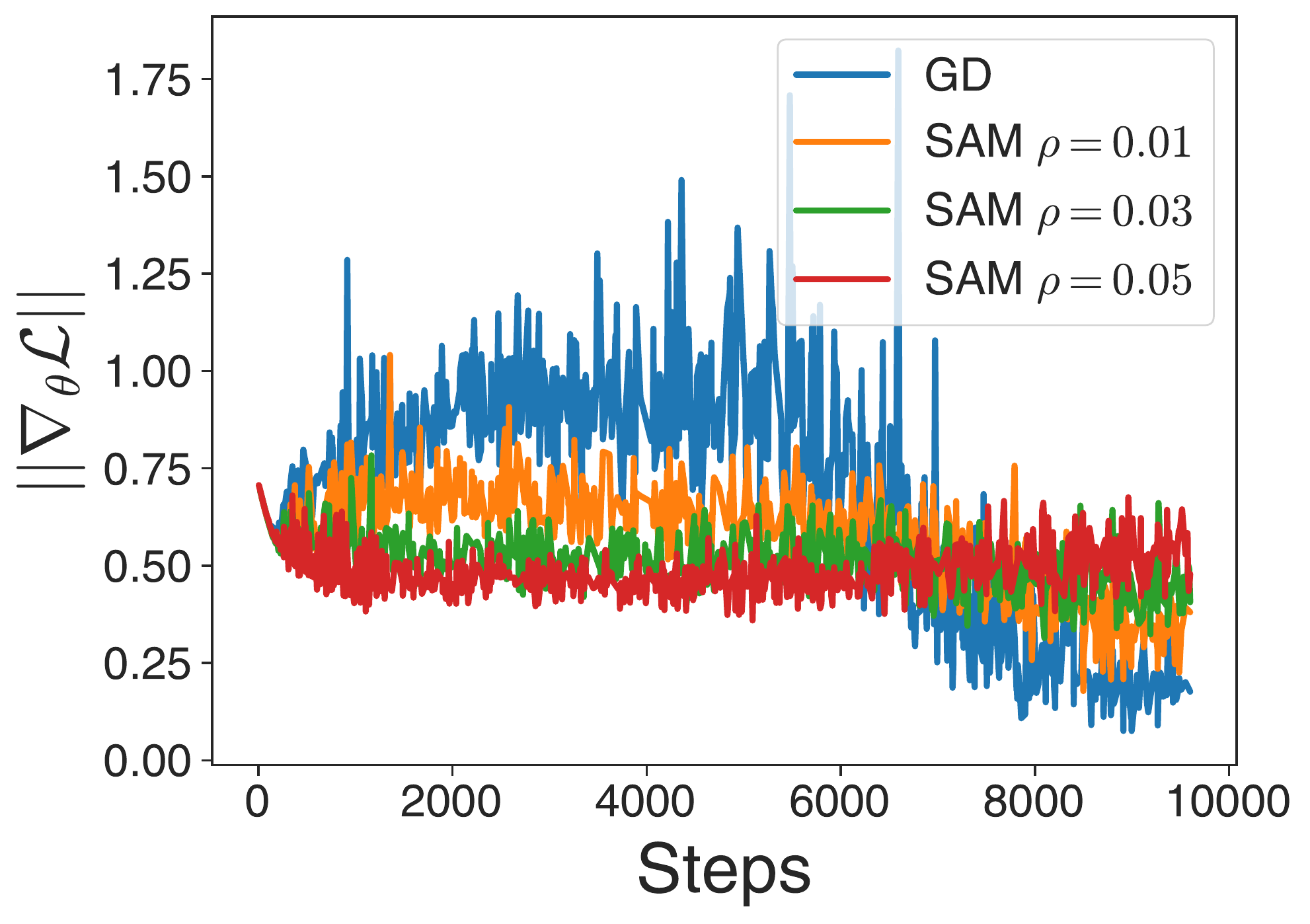}
    \caption{Minibatch gradient magnitudes for CIFAR-10 model trained on MSE loss. Magnitudes
    are steady early on in SGD training, but decrease at the end of training. Eigenvalue variation
    is smaller for increasing sam radius $\rad$.}
    \label{fig:mb_grad_mse}
\end{figure}

{\bf Minibatch gradient norm varies little.}\quad Another quantity of interest is the magnitude of the mini-batch gradients. For SGD, the
gradient magnitudes were steady during the first half of training and dropped by a factor
of $4$ at late times
(Figure \ref{fig:mb_grad_mse}). Gradient magnitudes were very stable for SAM, particularly for larger
$\rad$. This suggests that in practice, there may not be much
difference between the normalized and un-normalized SAM algorithms.
This is consistent with previous work which
showed that the generalization of the two approaches is similar
\cite{andriushchenko_understanding_2022}.

\section{Discussion}

\subsection{SAM as a dynamical phenomenon}

Much like the study of EOS before it, our analysis of SAM suggests that
sharpness dynamics near minima are insufficient to capture relevant phenomenology.
Our analysis of the quadratic regression model suggests that SAM already
regularizes the large eigenmodes at early times, and the EOS analysis shows how
SAM can have strong effects even in the large-batch setting. Our theory also
suggested that SGD has additional mechanisms to control curvature early on in
training as compared to full batch gradient descent.

The SAM schedule experiments provided further evidence that multiple phases of the optimization
trajectory are important for understanding the relationship between SAM and generalization.
If the important effect was the convergence to a particular minimum, then only late SAM
would improve generalization. If instead some form of ``basin selection'' was key,
then only early SAM would improve generalization. The fact that both are important \cite{andriushchenko_understanding_2022}
suggests that the entire optimization trajectory matters.

We note that while EOS effects are \emph{necessary} to understand some aspects of
SAM, they are certainly not \emph{sufficient}. As shown in Appendix \ref{app:quad_model_numerics}, the details of the behavior near the EOS have
a complex dependence on $\rad$ (and the model).
Later on in learning, especially
with a loss like cross entropy, the largest eigenvalues may decrease even without
SAM \cite{cohen_gradient_2022} - potentially leading the dynamics away from the EOS.
Small batch size may add other effects, and EOS effects become harder to understand
if multiple eigenvalues are at the EOS. Nonetheless, even in more complicated cases
the SAM EOS gives a good approximation to the control SAM has on the eigenvalues,
particularly at early times.

\subsection{Optimization and regularization are deeply linked}

% \aga{this is kind of more philosophical - thoughts on how to make this more concrete? I feel
% like reviewers might not like this part as much, but to me it's a key lesson from
% the work.}

This work provides additional evidence that understanding some regularization
methods may in fact require analysis of the optimization dynamics - especially
those at early or intermediate times. This is in contrast to approaches which seek
to understand learning by characterizing minima, or analyzing behavior
near convergence only.
A similar phenomenology has been observed in evolutionary dynamics - the basic
$0$th order optimization method - where the details of optimization trajectories
are often more important than the statistics of the minima to understand long-term
dynamics \cite{nowak_analysis_2015, park_dexceedance_2016, agarwala_adaptive_2019a}.

\section{Future work}

Our main theoretical analysis focused on the dynamics $\z$ and $\J$ under squared loss;
additional complications arise for non-squared losses like cross-entropy. Providing a
detailed quantitative characterization of the EOS dynamics under these more general
conditions is an important next step.
% \aga{Should I point this out explicitly? I feel like reviewers are already going to ask
% about this so should we just pre-empt them?} \ynd{Yes}

Another important open question is the analysis of SAM (and the EOS effect
more generally) under SGD. 
While Theorem \ref{thm:zj_dyn_sgd} provides some insight
into the differences, a full understanding would require an analysis of
$\expect_{\pmat}[(\z\cdot\v_{i})^{2}]$ for the different eigenmodes $\v_{i}$ -
which has only recently been analyzed for a quadratic loss function \cite{paquette_sgd_2021, paquette_homogenization_2022, paquette_implicit_2022, lee_trajectory_2022}.
Our analysis of the CIFAR10 models showed that the SGD gradient magnitude does not change
much over training. Further characterization of the SGD gradient statistics will also
be useful in understanding the interaction of SAM and SGD.

More detailed theoretical and experimental analysis of more complex settings may allow
for improvements to the SAM algorithm and its implementation in practice. A more detailed
theoretical understanding could lead to proposals for $\rad$-schedules, or improvements
to the core algorithm itself - already a
field of active research \cite{zhuang_surrogate_2022}.

Finally, our work focuses on optimization and training dynamics; linking these properties
to generalization remains a key goal of any further research into SAM and other
optimization methods.


\bibliography{sam_far_icml,second}
\bibliographystyle{icml2023}

%%%%%%%%%%%%%%%%%%%%%%%%%%%%%%%%%%%%%%%%%%%%%%%%%%%%%%%%%%%%%%%%%%%%%%%%%%%%%%%
%%%%%%%%%%%%%%%%%%%%%%%%%%%%%%%%%%%%%%%%%%%%%%%%%%%%%%%%%%%%%%%%%%%%%%%%%%%%%%%
% APPENDIX
%%%%%%%%%%%%%%%%%%%%%%%%%%%%%%%%%%%%%%%%%%%%%%%%%%%%%%%%%%%%%%%%%%%%%%%%%%%%%%%
%%%%%%%%%%%%%%%%%%%%%%%%%%%%%%%%%%%%%%%%%%%%%%%%%%%%%%%%%%%%%%%%%%%%%%%%%%%%%%%
\newpage
\appendix
\onecolumn

\section{Quadratic regression model}

\subsection{Rescaled dynamics}

\label{app:rescaled_dynamics}

The learning rate can be rescaled out of the quadratic regression model.
In previous work, the the rescaling
\begin{equation}
\tz = \lr\z,~\tJ = \lr^{1/2}\J
\end{equation}
which gave a universal representation of the dynamics.
The same rescaling in the SAM case gives us:
\begin{equation}
\begin{split}
\tz_{t+1}-\tz_{t} & = -(\tJ_{t}\tJ_{t}^{\top}+\rat (\tJ_{t}\tJ_{t}^{\top})^{2})\tz_{t}-\rat [(1+\rat\tJ_{t}\tJ_{t}^{\top})\tz_{t}]^{\top}\Q(\tJ_{t}^{\top}\tz_{t},\tJ_{t}^{\top}\cdot) \\
& +\frac{1}{2} \Q[\tJ_{t}^{\top}(1+\rat\tJ_{t}\tJ_{t}^{\top})\tz_{t},\tJ_{t}^{\top}(1+\rat\tJ_{t}\tJ_{t}^{\top})\tz_{t}]+O(||\tz^{3}||)
\end{split}
\end{equation}
\begin{equation}
\begin{split}
\tJ_{t+1} -\tJ_{t} & = - \Q(\tJ_{t}^{\top}(1+\rat\tJ_{t}\tJ_{t}^{\top})\tz_{t}, \cdot)-\rat\Q([(1+\rat\tJ_{t}\tJ_{t}^{\top})\tz_{t}]^{\top}\Q(\tJ_{t}^{\top}\tz_{t},\cdot) , \cdot)\\
& -\frac{1}{2} \rat^2\Q\left[\tJ_{t}^{\top}\Q(\tJ_{t}^{\top}\tz_{t}, \tJ_{t}^{\top}\tz_{t}), \cdot\right]+O(||\tz^3||)
\end{split}
\end{equation}
where $\rat$ is the rescaled SAM radius $\rat  = \rad/\lr$.

This suggests that, at least for gradient descent, the \emph{ratio} of the SAM radius
to the learning rate determines the amount of regularization which SAM provides.

\subsection{Average values, one step SGD}

\label{app:quad_average}

% \aga{Try to fix theorem formatting.}

We will prove Theorem \ref{thm:zj_dyn_sgd} first, and then derive Theorem
\ref{thm:j_dyn_gd} is as a special case.

\begin{reptheorem}{thm:zj_dyn_sgd}[]
Consider a second-order regression model, with $\Q$ initialized randomly with
i.i.d. components with $0$ mean and variance $1$.
For a model trained with SGD, sampling $\B$ datapoints independently
at each step, the change in $\z$ and $\J$ at the first step, averaged
over $\Q$ and the sampling matrix $\pmat_{t}$, is given by
\begin{equation}
\expect[\z_{1}-\z_{0}]_{\Q,\pmat}   = -\lr\bfr\J_{0}\J_{0}^{\top}(1+\rad[\bfr(\J_{0}\J_{0}^{\top})
+(1-\bfr)\diag(\J_{0}\J_{0}^{\top})])\z_{0}+O(\lr^2||\z||^{2})+O(\D^{-1})
\end{equation}
\begin{equation}
\expect_{\Q,\pmat}[\J_{1} -\J_{0}] = -\rad\lr\P (\bfr^{2}\z_{0}\z_{0}^{\top}
+\bfr(1-\bfr)\diag(\z_{0}\z_{0}^{\top})) \J_{0}+O(\rad^2\lr^2||\z||^2)+O(\lr^3||\z||^{3})
\end{equation}
where $\bfr \equiv\B/\D$ is the batch fraction.
\end{reptheorem}

\begin{proof}
We can write the SGD dynamics of the quadratic regression model as:
\begin{equation}
\z_{t+1}-\z_{t} = -\lr\J_{t} \J_{t}^{\top}\pmat_{t}\z_{t}  +\frac{1}{2}\lr^2 \Q(\J_{t}^{\top}\pmat\z_{t},\J_{t}^{\top}\pmat\z_{t})
\label{eq:SGD_in_z_general}
\end{equation}
\begin{equation}
\J_{t+1} -\J_{t} = -\lr \Q(\J_{t}^{\top}\pmat_{t}\z_{t}, \cdot)\,.
\label{eq:SGD_in_J_general}
\end{equation}
where $\pmat_{t}$ is a projection matrix which chooses the batch. It is
a $\D\times\D$ diagonal matrix with exactly $\B$ random $1$s on the diagonal,
with all other entries $0$. This corresponds to choosing $\B$ random elements,
without replacement, at each timestep.

For SAM with SGD, the SAM step is replaced with $\rad\J_{t}\pmat_{t}\z_{t}$ as well.
Expanding to lowest order, we have:
\begin{equation}
\z_{t+1}-\z_{t}  = -\lr(\J_{t}\J_{t}^{\top}+\rad (\J_{t}\J_{t}^{\top})\pmat_{t} (\J_{t}\J_{t}^{\top}))\pmat_{t}\z_{t}+O(||\z||^{2})\end{equation}
\begin{equation}
\begin{split}
\J_{t+1} -\J_{t} & = - \lr\Q(\J_{t}^{\top}(1+\rad\pmat_{t}\J_{t}\J_{t}^{\top})\pmat_{t}\z_{t}, \cdot)-\rad\lr\Q([\pmat_{t}\z_{t}]^{\top}\Q(\J_{t}^{\top}\pmat_{t}\z_{t},\cdot) , \cdot)\\
& +O(\rad^2\lr^2||\z||^2)+O(\lr^3||\z||^{3})
\end{split}
\end{equation}

Consider the dynamics of $\z$. Taking the average over $\pmat_{t}$,
we note that $\expect[\pmat] = \bfr\m{I}$. For any
fixed $\D\times\D$ matrix $\m{M}$, we also have:
\begin{equation}
\expect[\pmat_{t}\m{M}\pmat_{t}] = \bfr^{2}\m{M}+\bfr(1-\bfr)\diag(\m{M})+O(\D^{-1})
\end{equation}
Substituting, we have:
\begin{equation}
\expect_{\pmat_{t}}[\z_{t+1}-\z_{t}]   = -\lr\bfr\J_{t}\J_{t}^{\top}(1+\rad[\bfr(\J_{t}\J_{t}^{\top})+(1-\bfr)\diag(\J_{t}\J_{t}^{\top})])\z_{t}+O(||\z||^{2})+O(\D^{-1})
\end{equation}
Now consider the dynamics of $\J$. First we averaging over random initial $\Q$. At the
first step we have:
\begin{equation}
\expect_{\Q}[\J_{1} -\J_{0}]_{\al i}  = -\rad\lr \expect[\Q_{\al i j}(\pmat\z)_{\bt}\Q_{\bt j k}\J_{\gm k}(\pmat\z)_{\gm} ]+O(\rad^2\lr^2||\z||^2)+O(\lr^3||\z||^{3})
\end{equation}
\begin{equation}
\expect_{\Q}[\J_{1} -\J_{0}]_{\al i}  = -\rad\lr\P(\pmat\z)_{\al}\J_{\gm i}(\pmat\z)_{\gm}+O(\rad^2\lr^2||\z||^2)+O(\lr^3||\z||^{3})
\end{equation}
Averaging over $\pmat$ as well, we have:
\begin{equation}
\expect_{\Q,\pmat}[\J_{1} -\J_{0}] = -\rad\lr\P (\bfr^{2}\z\z^{\top}+\bfr(1-\bfr)\diag(\z\z^{\top})) \J+O(\rad^2\lr^2||\z||^2)+O(\lr^3||\z||^{3})+O(\D^{-1})
\end{equation}
\end{proof}

Theorem \ref{thm:j_dyn_gd} can be derived by first setting $\bfr = 1$. Given a singular
value $\sing_{\al}$ corresponding to singular vectors $\w_{al}$ and $\v_{\al}$ we have
$\sing_{\al} = \w_{\al}^{\top}\J\v_{\al}$. For small learning rates, the singular
value of $\J_{1}$ can be written in terms of the SVD of $\J_{0}$ as
\begin{equation}
\sing_{\al}(\J_{1}) = \w_{\al}(\J_{0})^{\top}\J_{1}\v_{\al}(\J_{0})+O(\lr^2)
\end{equation}
Therefore we can write
\begin{equation}
\sing_{\al}(\J_{1})-\sing_{\al}(\J_{0}) = \w_{\al}(\J_{0})^{\top}(\J_{1}-\J_{0})\v_{\al}(\J_{0})+O(\lr^2)
\end{equation}
Averaging over $\Q$ and $\pmat$ completes the theorem.

\subsection{Numerical results}

\label{app:quad_model_numerics}

The numerical results in Figure \ref{fig:sam_quad_model} were obtained by training the
model defined by the update Equation \ref{eq:unnorm_sam} in $\z$ and $\J$ space directly.
The tensor $\Q$ was randomly initialized with i.i.d. Gaussian elements at initialization,
and $\z$ and $\J$ were randomly initialized as well following the approach in 
\cite{agarwala_secondorder_2022}.
The figures correspond to $5$ independent initializations with the same statistics for
$\Q$, $\z$, and $\J$. All plots used $\D = 200$ datapoints with $\P = 400$ parameters.

For small $\lr$, the loss converges exponentially to $0$. In particular, the projection onto
the largest eigenmode decreases quickly %\aga{add figure?}
, which by the analysis of
Theorem \ref{thm:j_dyn_gd} suggests that SAM has only a small effect on the largest eigenvalues.

For larger $\lr$, by increasing $\rad$ the SAM dynamics seems to access the edge of stability
regime, where non-linear effects can stabilize the large eigenvalues of the curvature.
One way the original edge of stability dynamics was explored was to examine trajectories
at different learning rates \cite{cohen_gradient_2022}.
At small learning rate, training loss decreases
monotonically; at intermediate learning rates, the edge of stability behavior causes
non-monotonic but still stable learning trajectories, and finally, at large learning rate the
training is unstbale.

\begin{figure}[h]
    \centering
    \begin{tabular}{cc}
    \includegraphics[width=0.4\linewidth]{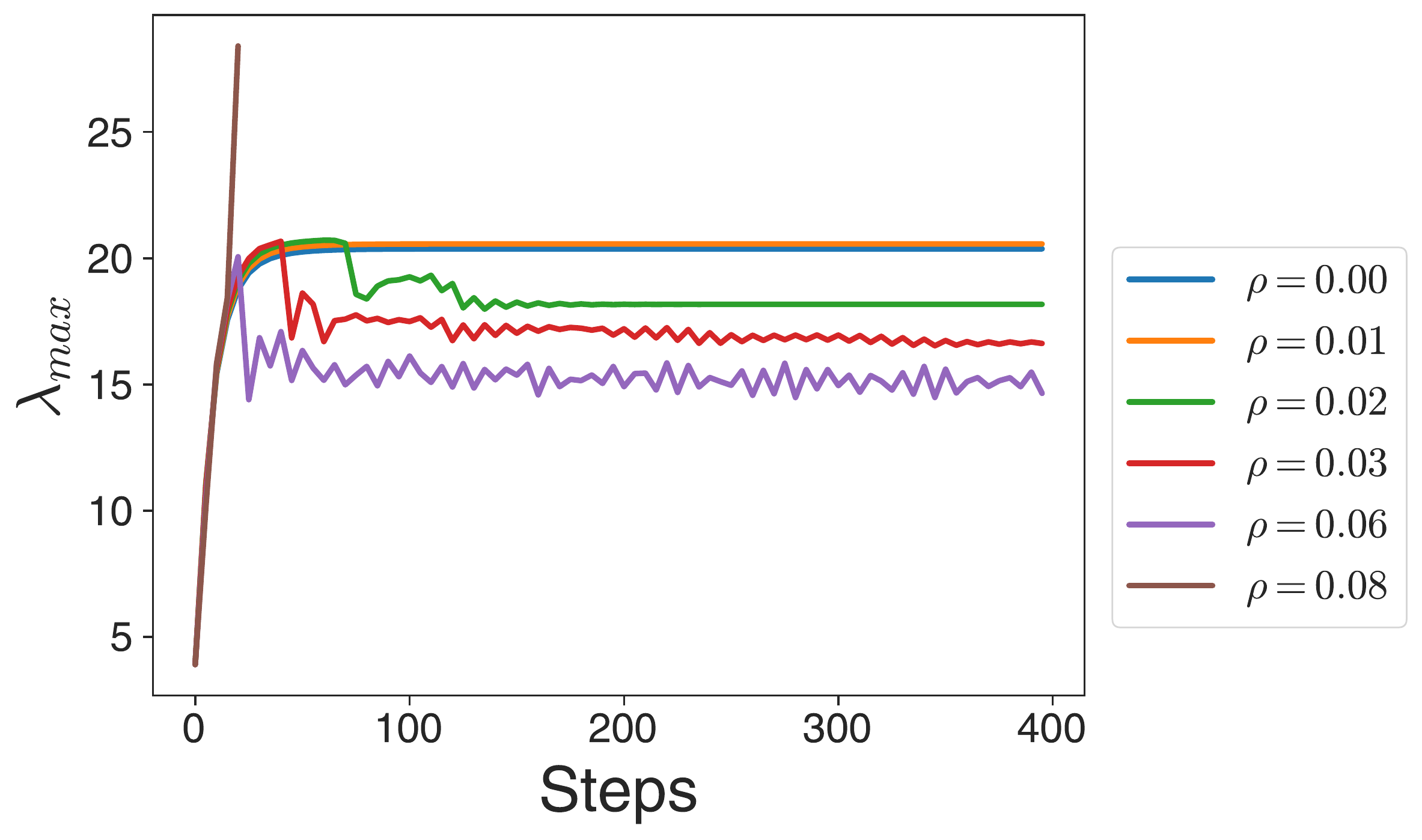} &
    \includegraphics[width=0.4\linewidth]{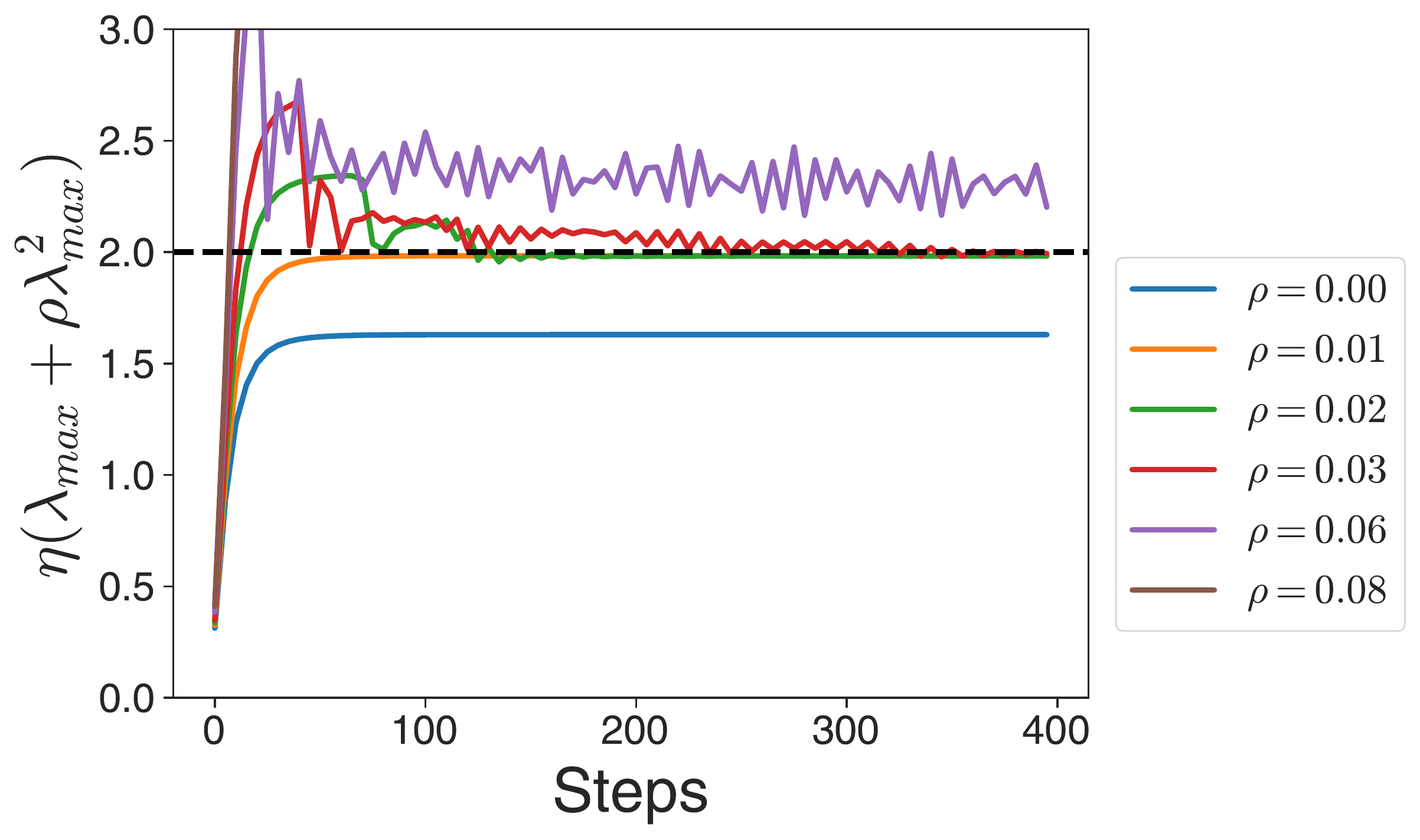}
    \end{tabular}
    \caption{For fixed $\lr$, as $\rad$ increases the largest eigenvalue of $\J\J^{\top}$
    decreases, until training is no longer stable (left). For intermediate $\rad$,
    the eigenvalue is very well predicted by $\lr(\lam_{max}+\rad\lam_{max}^2)= 2$ (right); however
    there is also a range of $\rad$ where training is still stable but $\lr(\lam_{max}+\rad\lam_{max}^2) > 2$ (purple curve).}
    \label{fig:quad_rad_sweep}
\end{figure}

We can similarly increase the SAM radius $\rad$ for fixed learning rate, and see an analogous
transition. If we pick $\lr$ such that the trajectory doesn't reach the non-linear edge of
stability regime, and increase $\rad$, we see that SAM eventually leads to a decrease in the
largest eigenvalues, before leading to unstable behavior (Figure \ref{fig:quad_rad_sweep}, 
left). If we plot $\lr(\lam_{max}+\rad\lam_{max}^2)$, we see that this normalized, effective
eigenvalue stabilizes very close to $2$ for a range of $\rad$, and for larger $\rad$ stabilizes
near but not exactly at $2$ (Figure \ref{fig:quad_rad_sweep}, right).
We leave a more detailed understanding of this stabilization for future work.

\section{SAM edge of stability}

\subsection{Proof of Theorem \ref{thm:eos_SAM}}

\label{app:sam_eos_proof}

We prove the following theorem, which gives us an alternate edge of stability
for SAM:

\begin{reptheorem}{thm:eos_SAM}[]
Consider a $\mathcal{C}^{\infty}$ model $\f(\th)$
trained using Equation \ref{eq:unnorm_sam} with MSE loss. Suppose that there
exists a point $\th^*$ where $\z(\th^*) = 0$. Suppose that for some
$\eps>0$, we have the lower bound
$\eps < \lr\lam_{i}(1+\rad\lam_{i})$ for the eigenvalues of the
positive-definite symmetric matrix
$\J(\th^*)\J(\th^*)^{\top}$. Given bounds on the largest eigenvalues,
there are two regimes:

\textbf{Convergent regime.} If $\lr\lam_{i}(1+\rad\lam_{i})<2-\eps$ for all
for all eigenvalues $\lam_{i}$ of $\J(\th^*)\J(\th^*)^{\top}$, there exists a neighborhood
$U$ of $\th^*$ such that $\lim_{t\to\infty}\z_{t} = 0$ with exponential convergence for
any trajectory initialized at $\th_{0}\in U$.

% \textbf{Divergent regime.} If $\lr\lam_{i}(1+\rad\lam_{i})>2+\eps$ for some eigenvector
% $\v_{i}$ of $\J(\th^*)\J(\th^*)^{\top}$, then there exists some $(q_{min}, d_{min})$
% such that for any $q<q_{min}$, $d<d_{min}$, given $S_{q, d}$, the connected neighborhood
% of $\th^*$ such that $||\th-\th^*||<q$ and $||\z(\th)||<d$ for all $\th$ in $S_{q, d}$.
% there exists some initialization $\th_{0}\in S_{q, d}$ such that the trajectory
% $\{\th_{t}\}$ leaves $S_{q,d}$ at some finite time $t$.
\textbf{Divergent regime.} If $\lr\lam_{i}(1+\rad\lam_{i})>2+\eps$ for some eigenvector
$\v_{i}$ of $\J(\th^*)\J(\th^*)^{\top}$, then there exists some $q_{min}$
such that for any $q<q_{min}$, given $B_{q}(\th^*)$, the ball of radius $q$
around $\th^*$,
there exists some initialization $\th_{0}\in B_{q}(\th^*)$ such that the trajectory
$\{\th_{t}\}$ leaves $B_{q}(\th^*)$ at some time $t$.
\end{reptheorem}

\begin{proof}
The SAM update for MSE loss can be written as:
\begin{equation}
\th_{t+1}-\th_{t} = -\lr\J^{\top}(\th_{t}+\rad\J^{\top}_{t}\z_{t})\z(\th_{t}+\rad\J^{\top}_{t}\z_{t})
\end{equation}
We will use the differentiability of $f(\th)$ to Taylor expand the update step, and
divide it into a dominant linear piece which leads to convergence, and an higher order
term which can be safely ignored in the long term dynamics.

Since the model $f(\th)$ is analytic at $\th^*$, there is a neighborhood $U_{r}$
of $\th^*$ with the following properties: for $\th\in U_{r}$, $\z$ and $\J$ there exists
a radius $r$ such that
\begin{equation}
\z(\th+\Delta\th)-\z(\th) = \J\Delta\th+\frac{1}{2}\frac{\partial^{2}\z}{\partial\th\partial\th'}(\Delta\th, \Delta\th)+\ldots
\end{equation}
\begin{equation}
\J(\th+\Delta\th)-\J(\th) = \frac{\partial^{2}\z}{\partial\th\partial\th'}(\Delta\th, \cdot)+ \frac{1}{2}\frac{\partial^{3}\z}{\partial\th_{1}\partial\th_{2}\partial\th_{3}}(\Delta\th, \Delta\th, \cdot)+\ldots
\end{equation}
for all $||\Delta\th||<r$. The derivatives which define the power series are taken
at $\th$. By Taylor's theorem, there exists
some $b >0$ such that we have the bounds
\begin{equation}
||\z(\th+\Delta\th)-\z(\th) - \J\Delta\th||\leq b ||\Delta\th||^{2}
\end{equation}
\begin{equation}
||\J(\th+\Delta\th)-\J(\th) - \frac{\partial^{2}\z}{\partial\th\partial\th'}(\Delta\th, \cdot)||\leq b ||\Delta\th||^{2}
\end{equation}
for $||\Delta\th||<r$ uniformly over $U_{r}$.
% Since $||\z|| = 0$ at $\th^*$, and $||\z||^{2}$ is itself analytic at
% $\th^*$, we can further posit that there exists a neighborhood $U_{r, a}\subset U_{r}$
% about $\th^*$ such that $r>a||\z(\th)||$ on $U_{r, a}$.

In addition, since $\J(\th^*)\J(\th^*)^{\top}$ has $\eps < \lr\lam_{i}(1+\rad\lam_{i})$
for all eigenvalues $\lam_{i}$, there exists a neighborhood $V_{\eps,1/2}$ of $\th^*$
such that $\eps/2 < \lr\lam_{i}(1+\rad\lam_{i})$ for all eigenvalues $\lam_{i}$ of
$\J\J^{\top}$, as well as
$\lam_{max}$ of $\J\J^{\top}$ is bounded by $\lr\lam_{i}(1+\rad\lam_{i})< 2-\eps/2$
in the \emph{convergent} case, and $2\lam_{max}(\th^*)$ in the \emph{divergent} case
for any $\th\in V_{\eps,1/2}$.
Finally, for any $d >0$, there
exists a connected neighborhood $T_{d}$ of $\th$ given by the set of points where
$||\z|| < d$.

% Consider the (non-empty) neighborhood $X_{r,a,d} = T_{d}\cap U_{r,a}\cap V_{\eps, 1/2}$
% given by the intersection of these sets. To recap, points $\th$ in $X_{r,a,d}$ have the
% following properties:
Consider the (non-empty) neighborhood $X_{r,a,d} = T_{d}\cap U_{r}\cap V_{\eps, 1/2}$
given by the intersection of these sets. To recap, points $\th$ in $X_{r, d}$ have the
following properties:
\begin{itemize}
    \item $\z$ and $\J$ have power series representations around $\th$ with radius
    $r > 0$.
    \item The second-order and higher terms are bounded by $b||\Delta\th||^2$
    uniformly on $X_{r,d}$, independently of $d$.
    % uniformly on $X_{r,a,d}$, independently of $a$ and $d$.
    \item $||\z(\th)|| < d$.
    % \item $||\z(\th)|| < d$ and $a||\z(\th)||< r$.
    \item The eigenvalues of $\J(\th)\J(\th)^{\top}$ are bounded from below by
    $\eps/2 < \lr\lam_{i}(1+\rad\lam_{i})$ and above by $\lr\lam_{i}(1+\rad\lam_{i})< 2-\eps/2$ (convergent case) or $2\lam_{max}(\th^*)$ (divergent case).
\end{itemize}

We now proceed with analyzing the dynamics. If $||\rad\J_{t}\z_{t}||< r$, then
we have:
\begin{equation}
\th_{t+1}-\th_{t} = -\lr (\J_{t}^{\top}+\rad\J_{t}^{\top}\J_{t}\J_{t}^{\top})\z_{t}+O(b||\rad\J_{t}^{\top}\z_{t}||^2)
\end{equation}
We note that $||\rad\J_{t}\z_{t}||< A||\z_{t}||$ on $X_{r,d}$ for some constant
$A$ independent of $d$, since the singular values of $\J_{t}$ are bounded
uniformly from above. Therefore, if we choose $d < r/A$, the power series representation
exists for all $\th\in X_{r,d}$.

If $||\th_{t+1}-\th_{t}||<r$, then both  $\z(\th_{t+1})-\z(\th_{t})$ as well as
$\J(\th_{t+1})-\J(\th_{t})$ can be represented as power series
centered around $\th_{t}$. We can again use the uniform bound on the singular values of
$\J$, as well as the uniform bound on the error terms to choose $d$ small enough
such that $||\th_{t+1}-\th_{t}||<r$ always for $\th_{t}\in X_{r,d}$.

Therefore, for sufficiently small $d$, we have:
\begin{equation}
\z(\th_{t+1})-\z(\th_{t}) = 
\z_{t+1}-\z_{t} = -\lr\J_{t}\J_{t}^{\top}[(1+\rad\J_{t}\J_{t}^{\top})\z_{t}]+O(h||\z_{t}||^2)
\end{equation}
\begin{equation}
\J(\th_{t+1})-\J(\th_{t}) = -\lr\frac{\partial^{2}\z}{\partial\th\partial\th'}(\J_{t}^{\top}\z_{t}, \cdot)+ O(h||\z_{t}||^{2})
\end{equation}
for some constant $h$ independent of $d$.

We first analyze the dynamics in the convergent case. We first establish that $||\z||^{2}$
decreases exponentially at each step, and then confirm that the trajectory remains in
$X_{\eps,1/2}$. Consider a single step in the eigenbasis of $\J_{t}\J_{t}^{\top}$.
Let $\zz(i)$ be the projection $\v_{i}\cdot\z$ for eigenvector $\v_{i}$
corresponding to eigenvalue $\lam_{i}$. Then we have:
\begin{equation}
\zz(i)^{2}_{t+1}-\zz(i)^{2}_{t} = (-\lr\lam_{i}(1+\rad\lam_{i})\zz(i)_{t}+O(||\z_{t}||^{2}))([2-\lr\lam_{i}(1+\rad\lam_{i})]\zz(i)_{t}+O(||\z_{t}||^{2}))
\end{equation}
From our bounds, we have
\begin{equation}
\zz(i)^{2}_{t+1}-\zz(i)^{2}_{t} \leq -\frac{1}{2}\eps\zz(i)_{t}^{2}+c||\z_{t}||^{3}
\end{equation}
By uniformity of the Taylor approximation we again have that $c$
is uniform, independent of $a$ and $d$.
Summing over the eigenmodes, we have:
\begin{equation}
||\z_{t+1}||^{2}-||\z_{t}||^2 \leq -\frac{1}{2}\eps||\z_{t}||^2+\D c||\z_{t}||^{3}
\end{equation}

If we choose $d<\frac{\eps}{4c\D }$, then
we have
\begin{equation}
||\z_{t+1}||^{2}-||\z_{t}||^2 \leq -\frac{1}{4}\eps||\z_{t}||^2
\end{equation}
Therefore $||\z_{t+1}||^{2}$ decreases by a factor of at least $1-\eps/4$ each step.

In order to complete the proof over the convergent regime, we need to show that
$\J_{t}$ changes by a small enough amount that the upper and lower bounds on the
eigenvalues are still valid - that is, the trajectory remains in $X_{\eps, 1/2}$.
Suppose the dynamics begins with initial residuals $\z_{0}$, and remains within
$X_{\eps, 1/2}$ for $t$ steps. Consider the $t+1$th step. The total change in $\J$
can be bounded by:
\begin{equation}
||\J_{t+1}-\J_{0}|| \leq  B\sum_{t}||\z_{t}||+C\sum_{t}||\z_{t}||^{2}
\end{equation}
for some constants $B$ and $C$ independent of $d$. The first term comes
from a uniform upper bound on
$-\lr\frac{\partial^{2}\z}{\partial\th\partial\th'}(\J_{t}^{\top}\z_{t}, \cdot)$,
and the second term comes from the uniform upper bound on the higher order corrections
to changes in $\J$ for each step. Using the bound on $||\z_{t}||$, we have:
\begin{equation}
||\J_{t+1}-\J_{0}|| \leq  \frac{4(B+C)}{\eps}||\z_{0}||
\end{equation}
If the right hand side of the inequality is less than $\eps^{(1+\delta)/2}$, for any
$\delta>0$, then the change in the singular values is $o(\eps^{1/2})$, the change
in the eigenvalues of $\J\J^{\top}$ is $o(\eps)$, and the trajectory remains in $V_{\eps, 1/2}$ at time
$t+1$. Let $d \leq \frac{1}{4(B+C)} \eps^{(3+\delta)/2}$. Then,
$||\J_{t+1}-\J_{0}|| \leq \eps^{(1+\delta)/2}$ for all $t$.

Therefore the trajectory remains within $X_{r,d}$, and $||\z_{t}||$
converges exponentially to $0$,
for any $d$ sufficiently small. Therefore there is a
neighborhood of $\th^*$ where $||\z||$ converges exponentially to $0$.

Now we consider the divergent regime. We will show that we can find initializations with
arbitrarily small $||\z||$ and $||\th-\th^*||$ which eventually have increasing $||\z||$.

Since $\J\J^{\top}$ is full rank, there exists some $\th_{0}$ in any neighborhood
of $\th^*$ such that $|\v_{m}\cdot\z(\th_{0})| > 0$ where $\v_{m}$ is
the direction of the largest eigenvalue of $\J\J^{\top}$. Consider such a
$\th_{0}$ in $X_{r,d}$ (and therefore in $T_{d}$ as well. The change in the
magnitude of this component $m$ of $\z$ is bounded from below by
\begin{equation}
\zz(m)^{2}_{1}-\zz(m)^{2}_{0} \geq \frac{1}{2}\eps\zz(m)_{1}^{2}-c||\z_{0}||^{3}
\end{equation}
Again the correction is uniformly bounded independent of $d$. Therefore the bound becomes
\begin{equation}
\zz(m)^{2}_{1}-\zz(m)^{2}_{0} \geq \frac{1}{4}\eps\zz(m)_{0}^{2}
\end{equation}
Choose $d_{min}$ such that the above bound holds for $d< d_{min}$. Furthermore, choose
$q_{min}$ so that the ball $B_{q_{min}}(\th^*)\subset X_{r, d_{min}}$. Pick
an initialization $\th_{0}\in B_{q}(\th^*)$ for $q < q_{min}$.

After a single step, there are two possibilities. The first is that 
$\th_{1}$ is no
longer in $B_{q_{min}}(\th^*)$. In this case the trajectory has left
$B_{q}(\th^*)$ and the proof is complete.

The second is that $\th_{1}$ remains in $B_{q_{min}}(\th^*)$. In this case, $z(m)_{1}^{2}$
is bounded from below by $(1+1/4\eps)z(m)_{0}^{2}$. If the trajectory remains
in $B_{q_{min}}(\th^*)$ for $t$
steps, we have the bound:
\begin{equation}
||\z_{t}||^{2} \geq (1+1/4\eps)^{t}z(m)_{0}^{2}
\end{equation}
Therefore, at some finite time $t$, $||\z_{t}||^{2} \geq d$, and $\th$ leaves $X_{r, d_{min}}$.
Therefore it leaves $B_{q}(\th^*)$.
This is true for any $q<q_{min}$. This completes the proof for the divergent case.

% Define $S_{q,d}$ as the connected neighborhood of $\th^*$ such that $||\th-\th^*|| <q$
% and $||\z(\th)|| < d$.
% Consider an initial point $\th_{0}\in S_{q,d}$ with $q<q_{min}$ and $d < d_{min}$,
% and $|\v_{m}\cdot\z(\th_{0})| > 0$.
% Note that under these constraints $S_{q,d}\subset X_{r, d_{min}}$.

% After a single step, there are two possibilities. The first is that $\th_{1}$ is no
% longer in $S_{q, d}$. In this case the trajectory has left the set and the proof is
% complete.

% The second is that $\th_{1}$ remains in $S_{q, d}$. In this case, $z(m)_{1}^{2}$
% is bounded from below by $(1+1/4\eps)z(m)_{0}^{2}$. If the trajectory remains
% in $S_{q,d}$ for $t$
% steps, we have the bound:
% \begin{equation}
% ||\z_{t}||^{2} \geq (1+1/4\eps)^{t}z(m)_{0}^{2}
% \end{equation}
% Therefore, at some finite time $t$, $||\z_{t}||^{2} \geq d$, and $\th$ leaves $S_{q,d}$.
% This is true for any $q<q_{min}$ and $d<d_{min}$.
% This completes the proof for the divergent case.

\end{proof}

\section{CIFAR-10 experiment details}

\subsection{Cross-entropy loss}

\label{app:x_ent_loss}

\begin{figure}[h]
    \centering
    \begin{tabular}{ccc}
        \includegraphics[width=0.31\linewidth]{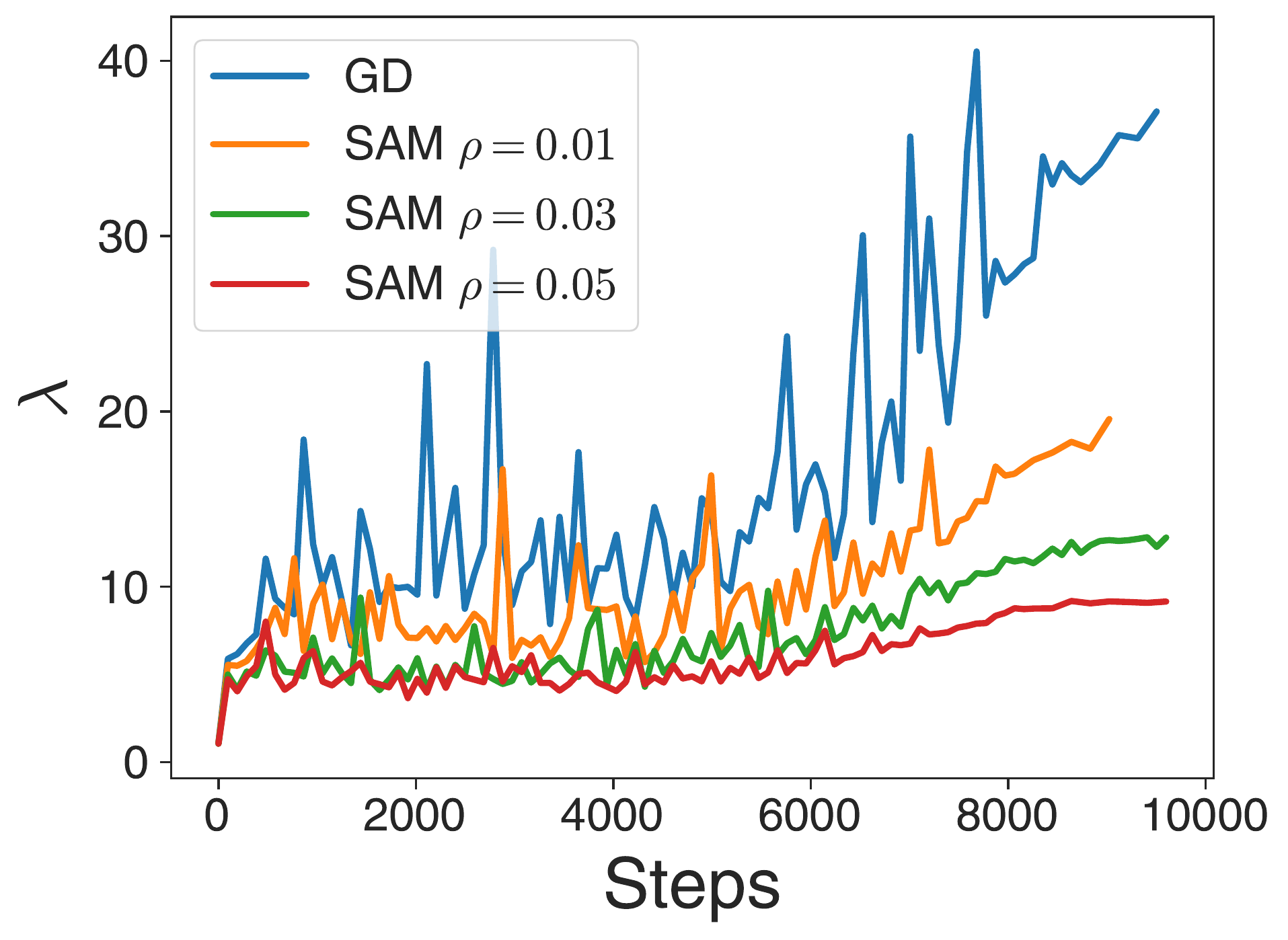}     & 
    \includegraphics[width=0.31\linewidth]{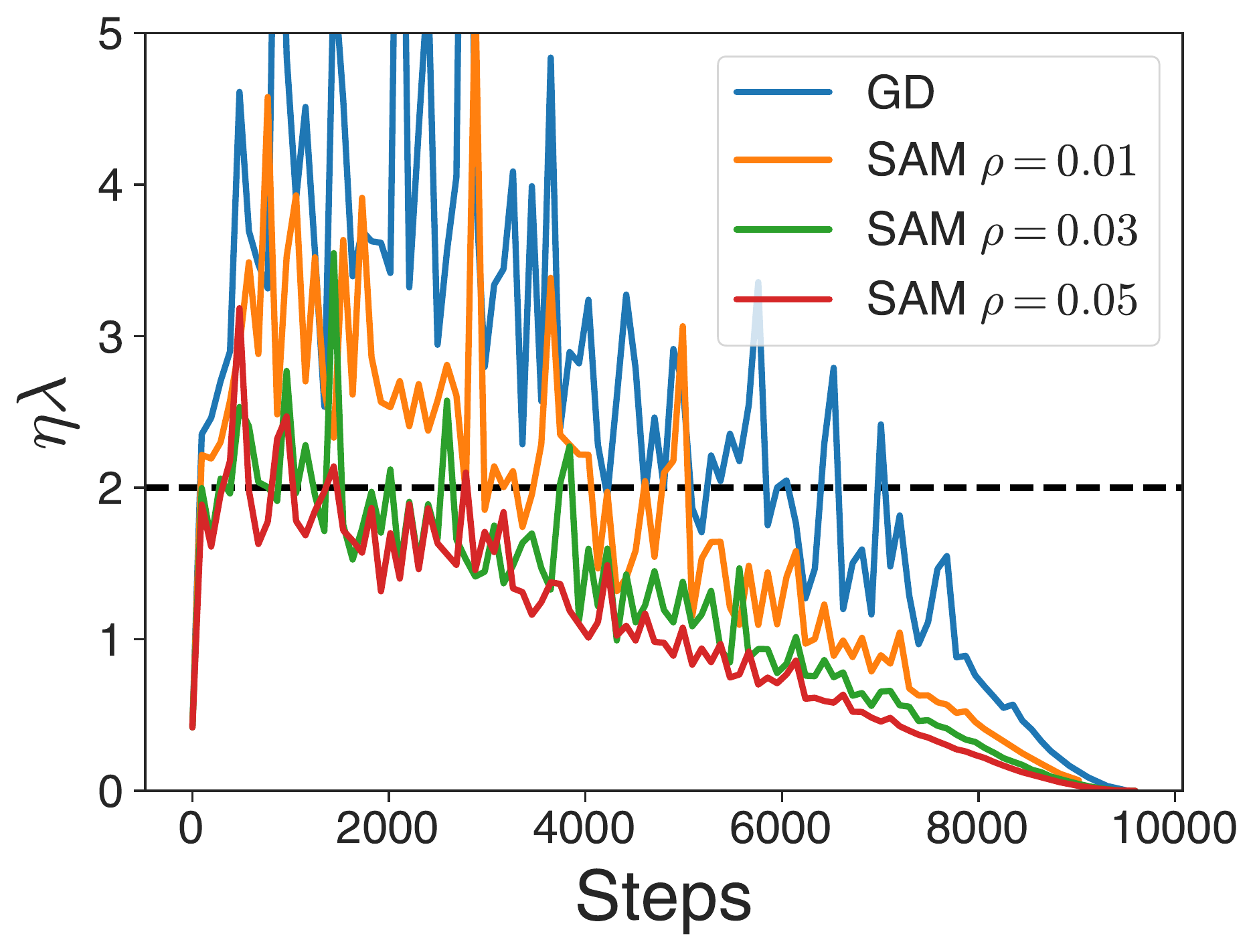}     & 
    \includegraphics[width=0.31\linewidth]{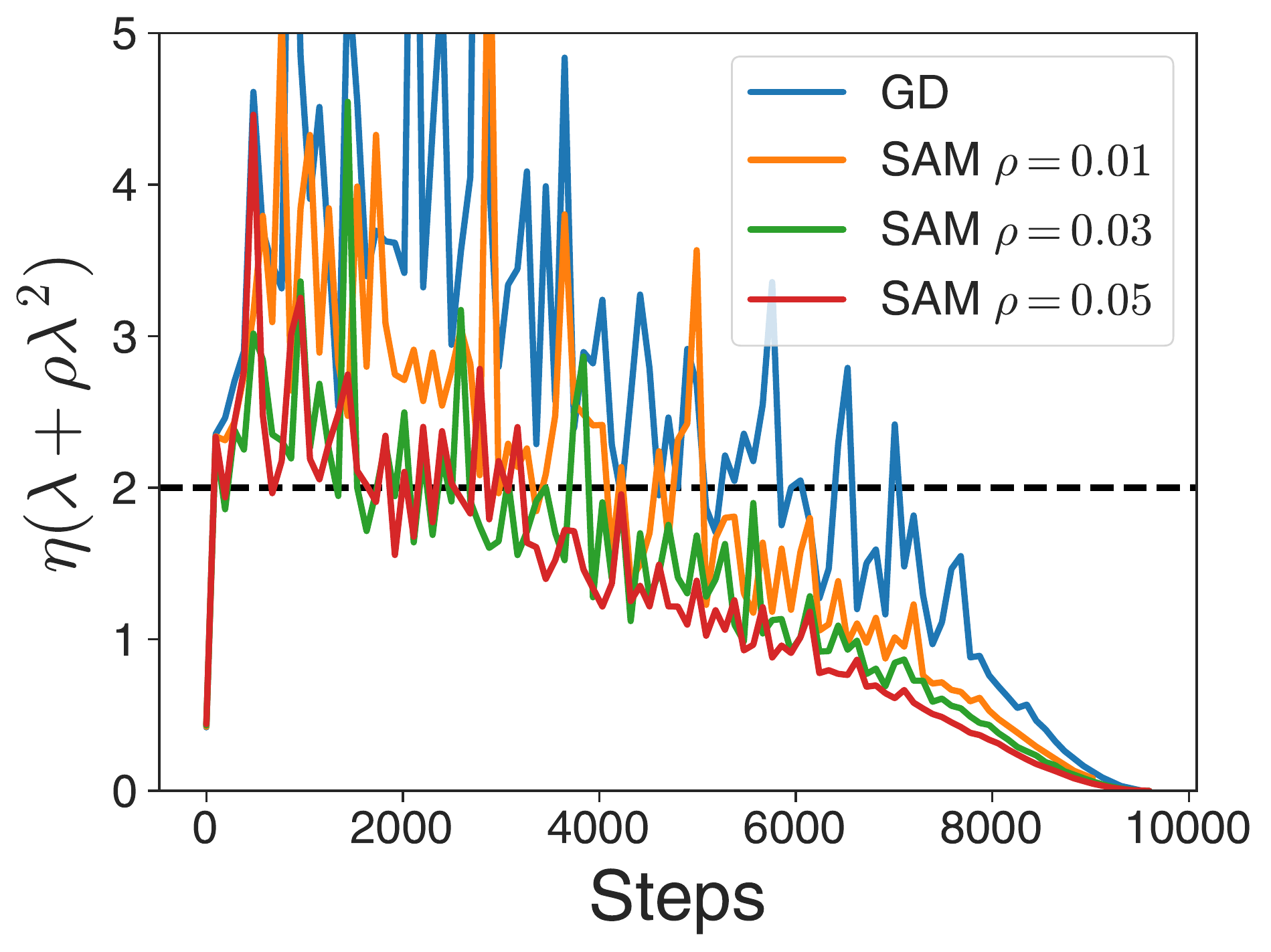}     
    \end{tabular}
    \caption{Largest Hessian eigenvalues for CIFAR10 trained with cross-entropy loss. Trends
    are similar to MSE loss (Figure \ref{fig:cifar10_mse_eos}), with the exception that normalized eigenvalues
    decrease from an earlier time.}
    \label{fig:cifar10_ce_eos}
\end{figure}

Many of the trends observed using MSE loss in Section \ref{sec:experiments} can also be observed
for cross-entropy loss. Eigenvalues generally increase at late times, and there is still
a regime where SGD shows EOS behavior in $\lr\lam_{max}$, while SAM shows EOS behavior in
$\lr(\lam_{max}+\rad\lam_{max}^2)$ (Figure \ref{fig:cifar10_ce_eos}). In addition, the
gradient norm is stil stable for much of training, with SGD gradient norm decreasing
at the end of training while SAM gradient norms stay relatively constant (Figure
\ref{fig:cifar10_ce_eos}).

There are also qualitative differences in the behavior. For example, the eigenvalue decrease
starts earlier in training. Decreasing eigenvalues for cross-entropy loss have been previously
observed \cite{cohen_gradient_2022}, and there is evidence that the origin of the effect is
due to the interaction of the logit magnitude with the softmax function. The gradient magnitudes
also have an initial rapid fall-off period. Overall more study is needed to understand how
the effects and mechanisms used by SAM depend on the loss used.

\begin{figure}[h]
    \centering
    \includegraphics[height=0.3\linewidth]{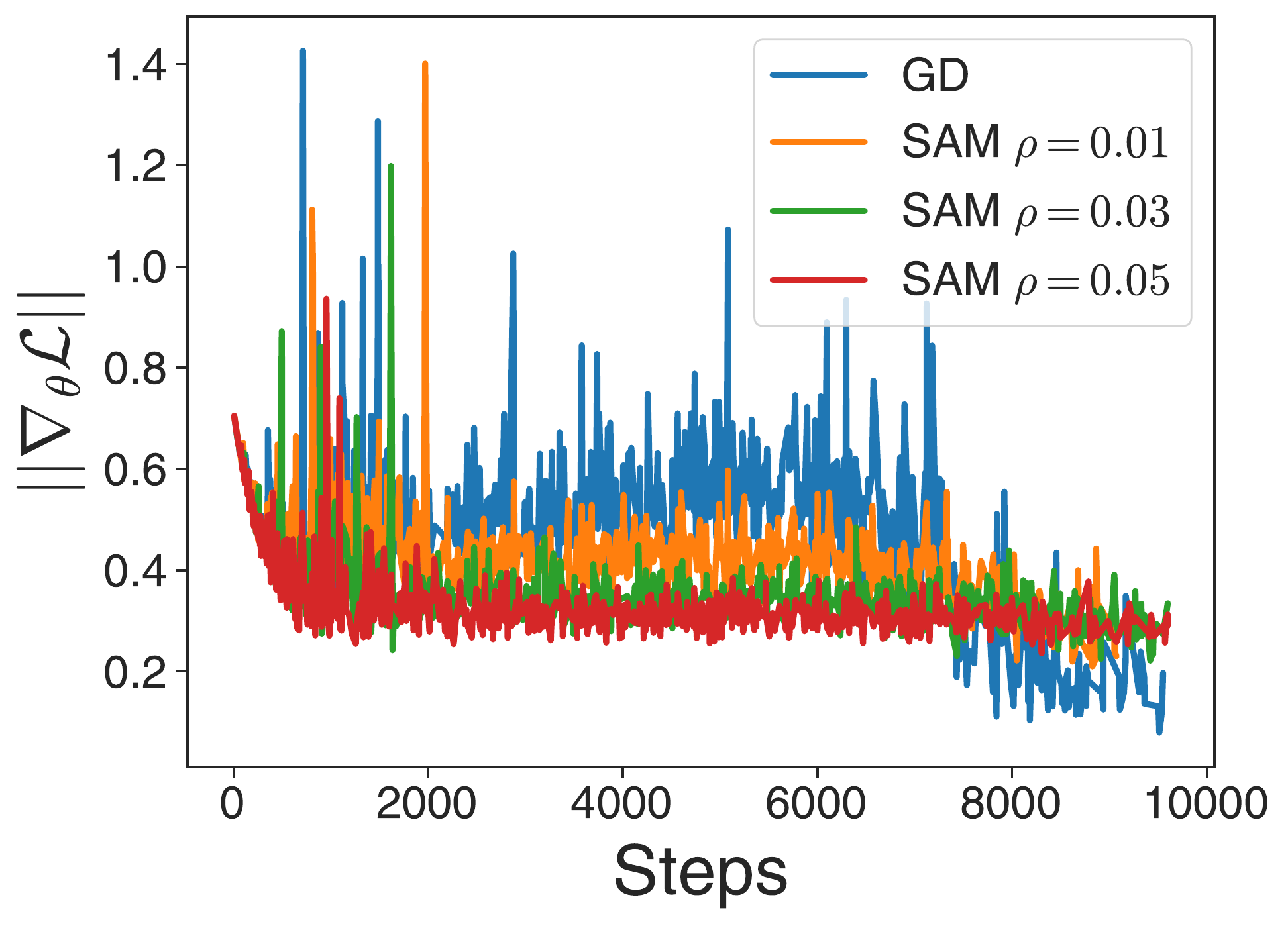}
    \caption{Minibatch gradient magnitudes for CIFAR-10 model trained on cross-entropy loss. 
    Trends are similar to MSE loss (Figure \ref{fig:cifar10_ce_eos}), with larger overall
    variation in gradient values.}
    \label{fig:mb_grad_ce}
\end{figure}

%%%%%%%%%%%%%%%%%%%%%%%%%%%%%%%%%%%%%%%%%%%%%%%%%%%%%%%%%%%%%%%%%%%%%%%%%%%%%%%
%%%%%%%%%%%%%%%%%%%%%%%%%%%%%%%%%%%%%%%%%%%%%%%%%%%%%%%%%%%%%%%%%%%%%%%%%%%%%%%

\end{document}

%% file: math_commands.tex
\usepackage{amsmath,amsfonts,bm}

\def\eqref#1{equation~\ref{#1}}
\def\1{\bm{1}}

\def\eps{{\epsilon}}

\DeclareMathAlphabet{\mathsfit}{\encodingdefault}{\sfdefault}{m}{sl}
\SetMathAlphabet{\mathsfit}{bold}{\encodingdefault}{\sfdefault}{bx}{n}
\newcommand{\tens}[1]{\bm{\mathsfit{#1}}}

\def\tJ{{\tens{J}}}

\newcommand{\lr}{\alpha}